\pdfoutput=1

\documentclass[11pt]{article}

\usepackage{naacl2021}

\usepackage{times}
\usepackage{latexsym}
\usepackage{graphicx}
\usepackage{xcolor}
\usepackage{amsfonts,amsmath}
\usepackage{array}
\usepackage{subcaption}
\usepackage{multirow}
\usepackage{changepage}
\usepackage{booktabs}
\usepackage{todonotes}
\usepackage[T1]{fontenc}

\usepackage[utf8]{inputenc}

\usepackage{microtype}

\title{Morph Call: Probing Morphosyntactic Content of Multilingual Transformers}

\author{Vladislav Mikhailov\textsuperscript{1,2},
~Oleg Serikov\textsuperscript{2,3},
~Ekaterina Artemova\textsuperscript{2,4} \\
  \textsuperscript{1} SberDevices, Sberbank, Moscow, Russia\\
  \textsuperscript{2} HSE University, Moscow, Russia\\
  \textsuperscript{3} Neural Networks and Deep Learning Lab\\
Moscow Institute of Physics and Technology, Dolgoprudny, Russia\\
  \textsuperscript{4} Huawei Noah's Ark lab, Moscow, Russia\\
  \tt{Mikhaylov.V.Nikola@sberbank.ru}  \tt{\{oserikov,elartemova\}@hse.ru}
\\}

\begin{document}
\maketitle
\begin{abstract}
The outstanding performance of transformer-based language models on a great variety of NLP and NLU tasks has stimulated interest in exploring their inner workings. Recent research has focused primarily on higher-level and complex linguistic phenomena such as syntax, semantics, world knowledge, and common sense. The majority of the studies are anglocentric, and little remains known regarding other languages, precisely their morphosyntactic properties. To this end, our work presents {\bf Morph Call}, a suite of 46 probing tasks for four Indo-European languages of different morphology: English, French, German and Russian. We propose a new type of probing task based on the detection of guided sentence perturbations. We use a combination of neuron-, layer- and representation-level introspection techniques to analyze the morphosyntactic content of four multilingual transformers, including their less explored distilled versions. Besides, we examine how fine-tuning for POS-tagging affects the model knowledge. The results show that fine-tuning can improve and decrease the probing performance and change how morphosyntactic knowledge is distributed across the model. The code and data are publicly available, and we hope to fill the gaps in the less studied aspect of transformers.
\end{abstract}

\section{Introduction}
In the last few years, transformer language models \citep{vaswani2017attention} have accelerated the growth in the field of NLP. The models have established new state-of-the-art results in multiple languages and even demonstrated superiority in NLU benchmarks compared to human solvers \citep{raffel2020exploring,xue2020mt5,he2020deberta}. Their distilled versions, or so-called student models, have shown competitive performance on many NLP tasks while having fewer parameters \citep{tsai2019small}. However, many questions remain on how these models work and what they know about language. The previous research focuses on what
knowledge has been learned during and after pre-training phases \citep{chiang2020pretrained,rogers2020primer}, and how it is affected by fine-tuning \citep{gauthier2019linking,peters2019tune,miaschi2020linguistic,merchant2020happens}. Besides, a wide variety of language phenomena has been investigated including syntax \citep{hewitt-manning-2019-structural, liu2019linguistic}, world knowledge \citep{petroni2019language,jiang2020can}, reasoning \citep{van2019does}, common sense understanding \citep{zhou2020evaluating,klein2019attention}, and semantics \citep{ettinger2020bert}. 

Most of these studies involve \emph{probing} which measures how well linguistic knowledge can be inferred from the intermediate representations of the model. The methods range from individual neuron analysis \citep{dalvi2020analyzing,durrani2020analyzing}, examination of attention mechanisms \citep{kovaleva-etal-2019-revealing,vig-belinkov-2019-analyzing}, correlation-based similarity measures \citep{wu2020similarity}, to probing tasks accompanied by linguistic supervision \citep{adi2016fine,conneau2018you}.

Despite growing interest in interpreting the models, morphology has remained understudied, specifically for languages other than English. The majority of prior works on this subject are devoted to the introspection of machine translation models, word-level embedding models, or transformers, fine-tuned for POS-tagging  (see Section \ref{related-work}). 

To this end, we introduce \textbf{Morph Call}, a probing suite for the exploration of morphosyntactic content in transformer language models. The contributions of this paper are summarized as follows. First, we propose 46 probing tasks in four Indo-European languages of different morphology:  Russian, French, English, and German. Inspired by techniques for model acceptability judgments \citep{warstadt-etal-2019-neural} and adversarial training \citep{alzantot2018generating,tan2020s,tan2020mind}, we present a new type of probing tasks based on the detection of guided sentence perturbations. Since the latter is automatically generated, the tasks can be adapted to other languages. Second, we use complementary probing methods to analyze four multilingual transformer encoders, including their distilled versions. We examine how fine-tuning for POS-tagging affects the probing performance and establish count-based and non-contextualized baselines for the tasks. Finally, we publicly release the tasks and code\footnote{\url{https://github.com/morphology-probing/morph-call}}, hoping to fill the gaps in the less studied aspect of transformers.

\section{Related Work}
\label{related-work}
A large body of recent research is devoted to analyzing and interpreting the linguistic capacities of pre-trained contextualized encoders. The most common approach is to train a simple classifier for solving a probing task over the word- or sentence-level features produced by the models \citep{conneau2018you,liu2019linguistic}. The classifier's performance is used as a proxy to assess the model knowledge about a particular linguistic property. However, lately, the method has been critiqued: is the property truly learned by the model, or does the model encode the property for the classifier to easily extract it given the supervision? Besides, a new set of additional classifier parameters can make it challenging to interpret the results \citep{hewitt2019designing,hewitt2019structural,saphra-lopez-2019-understanding,voita2020information}.

Nevertheless, the probing classifiers are widely applied in the field of model interpretation, including morphology. One of the first works on morphological content is carried out on machine translation models where the classifier is learned to predict POS-tags in multiple languages \citep{belinkov2017neural,belinkov2018evaluating}. The latest studies involving POS properties in transformers show that they are predominantly captured at the lower layers \citep{tenney2018what,liu-etal-2019-linguistic,rogers2020primer}, and can be evenly distributed across all layers \citep{durrani-etal-2020-analyzing}. \emph{Amnesic} probing explores how removing information at a particular layer affects the probe performance at the final layer \citep{elazar2020bert}. This allows measuring the layer importance with respect to a linguistic property. The results claim that removing POS information may affect the performance more at the higher layers as compared to the lower ones.

Another line of research is devoted to various linguistic phenomena at the juxtaposition of morphology, syntax, and semantics. LSTM-based models and transformers are probed to capture subject-verb agreement in different languages \citep{linzen2016assessing,giulianelli2018under,ravfogel2018can,goldberg2019assessing}. Recently, the agreement has been at the core of inflectional perturbations for adversarial training \citep{tan-etal-2020-morphin}, and linguistic acceptability judgments along with morphological, syntactic, and semantic violations \citep{warstadt2019neural}.

Our work is closely related to \citep{edmiston2020systematic} who explore morphological properties and subject-verb agreement in the hidden representations and self-attention heads of transformer models. However, there are several differences. First, we investigate the knowledge in multilingual transformers and their distilled versions instead of monolingual ones. Second, we carry out the experiments on an extended set of tasks, such as detecting syntactic and inflectional perturbations (see Section \ref{probing_tasks}). Third, we apply several probing methods to analyze from different perspectives. Finally, we study the impact of fine-tuning for POS-tagging on the probe performance. Despite the similarities and differences, we find the studies complementary.

Finally, such benchmarks as LINSPECTOR \citep{sahin-etal-2020-linspector} and XTREME \citep{hu2020xtreme} provide means for evaluation of multilingual embedding models and cross-lingual transferring methods with regards to multiple linguistic properties, specifically morphology.

\section{Method}
\label{method}
\subsection{Morphosyntactic Inventories}
\label{morph-inventories}
This paper investigates four Indo-European languages that fall under different morphological types: Russian, French, English, and German. Russian and French have fusional morphology, while English is an analytic language, and German exhibits peculiarities of fusional and agglutinative types. We consider the nominal morphosyntactic features of Number, Case, Person, and Gender. Even though the feature inventory is mostly shared across the languages, the latter differ significantly in their richness of morphology \citep{baerman2007syncretism}. The morphosyntactic inventories of the analyzed languages are outlined in Table \ref{features}.

\begin{table*}[ht!]
    \centering
	\begin{center}
	\scalebox{.82}{\begin{tabular}{c|c|c|c|c}
		\toprule 
		\textbf{Feature} \textbf{$\backslash$} \textbf{Language} & \textbf{English} & \textbf{French} & \textbf{German} & \textbf{Russian} \\
		\midrule
		\textbf{Number} & $\{Sing, Plur\}$ & $\{Sing, Plur\}$ & $\{Sing, Plur\}$ & $\{Sing, Plur\}$ \\
		\midrule
		\textbf{Case} & – & – & $\{Nom, Acc, Dat, Gen\} $ & $\{Nom, Acc, Dat, Gen, Loc, Ins \}$ \\ 
		\midrule 
		\textbf{Person} & $\{ 1, 2, 3 \}$ & $\{ 1, 2, 3 \}$ & $\{ 1, 2, 3 \}$ & $\{ 1, 2, 3 \}$ \\
		\midrule
		\textbf{Gender} & – & $\{ Masc, Fem \}$ & $\{ Masc, Fem, Neut \}$ & $\{Masc, Fem, Neut\}$ \\
		\bottomrule
	\end{tabular}}
	\caption{Analyzed languages and their morphosyntactic feature inventories.}
	\label{features}
    \end{center}
\end{table*}

\subsection{Probing Tasks}
\label{probing_tasks}
\paragraph{Data} We use sentences from the Universal Dependencies (UD) \citep{nivre2016universal} for all our probing tasks, keeping in mind possible inconsistency between the Treebanks \citep{de2017assessing,alzetta2017dangerous,droganova2018data}, and consequent inconsistency in dataset sizes across languages. All sentences are filtered by a 5-to-25 token range, and each task is split into 80/10/10 train/val/test partitions with no sentence overlap. The partitions are balanced by the number of instances per target class. Notably, the availability of the UD Treebanks in different languages allows for an adaptation of the method to the other ones. The used Treebanks are listed in Appendix \ref{app:Treebank_desc}, and a brief statistics of the tasks is presented in Appendix \ref{app:data_stata}.

\paragraph{Task Description} We construct four groups of probing tasks framed as binary or multi-class classification tasks: \textbf{Morphosyntactic Features}, \textbf{Masked Token}, \textbf{Morphosyntactic Values} and \textbf{Perturbations}.

\vspace{0.5em}\noindent \textbf{Morphosyntactic Features} probe the encoder for the occurrence of the morphosyntactic properties. The goal is to detect if a word exhibits a particular property based on its contextualized representation. Consider an example for the Russian sentence \textit{`The clock stopped in a month.'}:
\begin{align*}
\text{Chasy }
\underbrace{\text{\textbf{ostanovilis'}}}_{\text{to stop+3PL+PST (1)}}
\text{ }
\underbrace{\text{\textbf{cherez}}}_{\text{in (0)}}
\text{ mesyats .}
\end{align*}
Here, the target words are indicated by bold, and the labels denote if they have the category of Number.

\vspace{0.5em}\noindent \textbf{Masked Token} tasks are analogous to \textbf{Morphosyntactic Features} with the exception that the target word is replaced with a tokenizer-specific mask token. 
The tasks test if it is possible to recover the properties of the masked token purely from the context. Below is an example where the sentence mentioned above \textit{`The clock stopped in a month.'} contains masked target words, and labels denote the occurrence of the Number feature at the position of the token:

\begin{align*}
\text{Chasy }
\underset{1}{\textbf{[MASK]}}
\text{ cherez mesyats .}\\
\text{Chasy ostanovilis' }
\underset{0}{\textbf{[MASK]}}
\text{ mesyats .}
\end{align*}

\vspace{0.5em}\noindent \textbf{Morphosyntactic Values} is a group of k-way classification tasks for each feature where \emph{k} is the number of values that the feature can take (see Table \ref{features}). For instance, the goal is to identify whether the word \emph{girl} is in the singular or plural form: `The \textbf{girl} has either pink or brown.'

\vspace{0.5em}\noindent \textbf{Perturbations} tasks test the encoder sensitivity to various sentence perturbations. Removing words from a text has recently been used to obtain adversarial attacks \citep{liang2017deep,li2018textbugger}, whereas inflectional perturbations have been applied for adversarial training of transformers \citep{tan2020s,tan2020mind}. In contrast, we extend the perturbations to probe the encoders for linguistic knowledge. To this end, we construct eight tasks that involve syntactic perturbations and inflectional perturbations in the subject-predicate agreement and deictic words. Note that we apply a set of language-specific rules to control the quality of the error generation procedure. To obtain the inflectional candidates, we make use of pymorphy2 for Russian \citep{korobov2015morphological}, lemminflect\footnote{\url{https://github.com/bjascob/LemmInflect}} for English, and word paradigm tables from Wiktionary for French\footnote{\url{https://dumps.wikimedia.org/frwiktionary/latest/}} and German\footnote{\url{https://dumps.wikimedia.org/dewiktionary/latest/}}.

\textbf{Stop-words Removal} involves corruption of a syntax tree by removing stop-words. We use lists of stop-words provided by NLTK library \citep{loper2002nltk}. Consider an example of the French sentence \textit{`\textbf{Les} Irakiens \textbf{ont} tout détruit \textbf{à le} Koweit'}, where the bolded words correspond to the removed stop-words.

\textbf{Article Removal} is a special case of the previous task, revealing whether the encoders are sensitive to discarded articles. This task is only constructed for French, English, and German. Note that such perturbation may also strain the semantics of the sentence: \textit{`It's on loan, by \textbf{the} way'.}

\textbf{Subject Number} includes inflectional perturbations of the subject in the main clause with respect to the Number: \textit{`The \textbf{girls} has either pink or brown.'}

\textbf{Subject Case} comprises errors in the subject form of Case for Russian. Consider an example of the perturbed sentence \textit{Kak \textbf{vy} vidite situatsiyu v Rossii?} `How do you find the situation in Russia?', where the nominative form of the subject \emph{vy} `you' is changed to the accusative:
\begin{align*}
\text{Kak}
\underbrace{\text{\textbf{vas}}}_{\text{you+2PL+ACC}}
\text{vidite situatsiyu v Rossii ?}
\end{align*}

\textbf{Predicate Number} incorporates perturbations of the predicate in the main clause regarding the Number feature:
\textit{`It \textbf{make} a huge difference.'}

\textbf{Predicate Gender} contains errors in the Gender form of the predicate in the main clause. For example, the masculine form of the predicate \emph{byl} `was' in the Russian sentence \textit{Dosug \textbf{byl} ves'ma odnoobrazen} `The leisure was pretty monotonous' is changed to the feminine:
\begin{align*}
\text{Dosug}
\underbrace{\text{\textbf{byla}}}_{\text{to be+3SG+FEM}}
\text{ves'ma odnoobrazen .}
\end{align*}

\textbf{Predicate Person} comprises perturbations in the Person form of the predicate in the main clause. For instance, the Russian sentence \textit{Ya \textbf{poedu} v Moskvu} `I will go to Moscow' contains the perturbed predicate in the form of the second Person instead of the first one:
\begin{align*}
\text{Ya }
\underbrace{\text{\textbf{poedesh'}}}_{\text{to go+2SG}}
\text{ v Mosckvu .}
\end{align*}

\textbf{Deictic Word Number} involves perturbations generated by the inflection of demonstrative pronouns (only in English and German). For example, the singular form of the pronoun \emph{dieser} `this' is changed to the plural form \emph{diesen} `these' in the sentence \textit{Siehe zu \textbf{dieser} Technik auch} `See also this technique':
\begin{align*}
\text{Siehe zu }
\underbrace{\text{\textbf{diesen}}}_{\text{this+PL+DAT}}
\text{ Technik auch .}
\end{align*}

\section{Experimental Setup}
\label{experimental_setup}

\begin{figure*}[ht!]
  \centering
  \includegraphics[width=.95\textwidth]{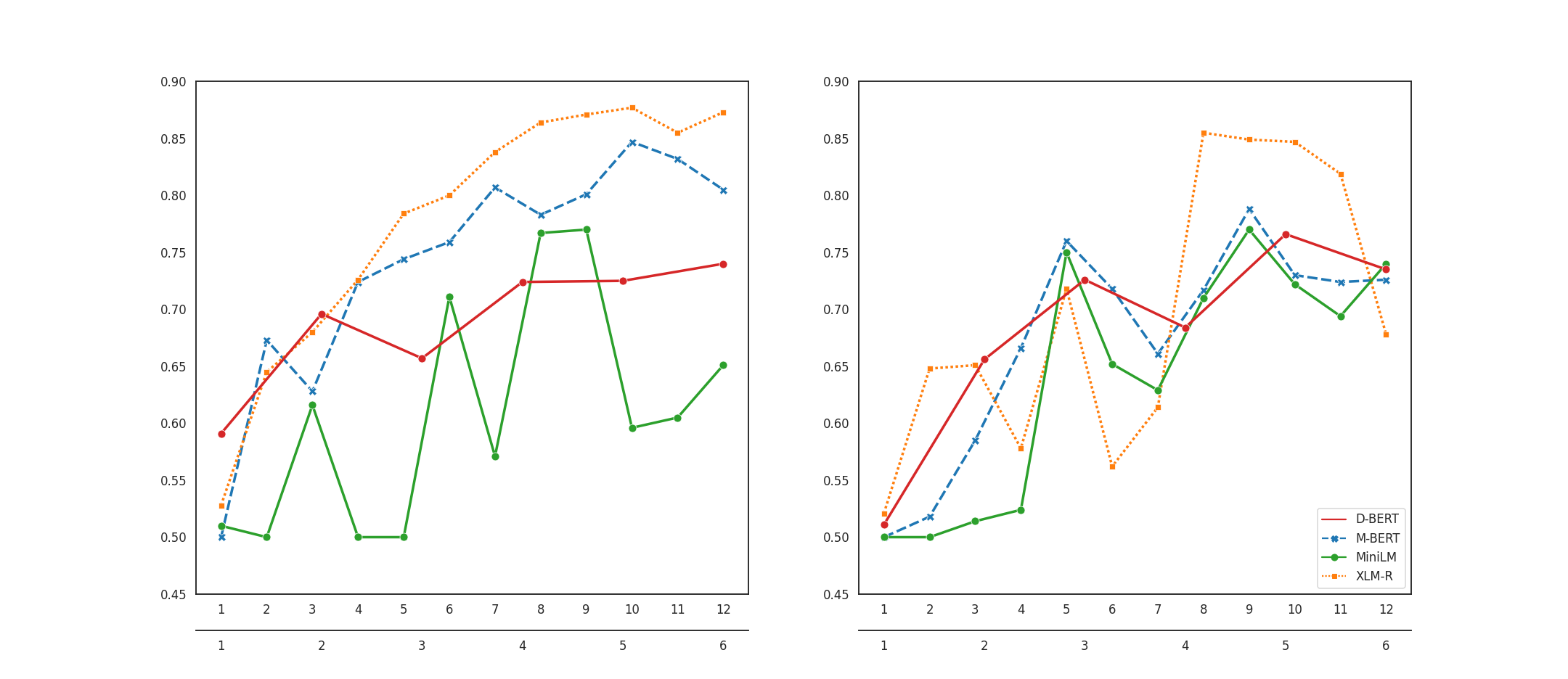}
  \caption{The performance of the probing classifier on \textbf{Case} masked token task for Russian. X-axis=Layer index score. Y-axis=Accuracy score. Left: pre-trained models. Right: fine-tuned models.}
  \label{mask_ru_case}
\end{figure*}

\subsection{Models}
\label{models}
The experiments are run on the following multilingual transformer models released as a part of HuggingFace library \citep{wolf2019huggingface}:

\vspace{0.5em}\noindent \textbf{M-BERT} \citep{devlin-etal-2019-bert} was pre-trained over concatenated monolingual Wikipedia corpora in 104 languages.

\vspace{0.5em}\noindent \textbf{D-BERT} \citep{sanh2019distilbert} or DistilBERT is a 6-layer distilled version of \textbf{M-BERT} model.

\vspace{0.5em}\noindent \textbf{XLM-R} \citep{conneau2019unsupervised} was pre-trained over filtered CommonCrawl data in 100 languages \citep{wenzek2019ccnet}.

\vspace{0.5em}\noindent \textbf{MiniLM} \citep{wang2020minilm} is a distilled \textbf{M-BERT} model that uses \textbf{XLM-R} tokenizer.

\vspace{0.5em}\noindent Each model under investigation has two instances for each language:
\begin{enumerate}
    \item \emph{Fine-tuned model} is a transformer model fine-tuned for POS-tagging. We use the UD Treebanks and HuggingFace library for fine-tuning. The data is randomly split into 80/10/10 train/val/test sets.
    \item \emph{Pre-trained model} is a non-tuned transformer model with frozen weights.
\end{enumerate}

\subsection{Probing Methods}
\label{subsec:probe_methods}
\paragraph{Probing Classifiers}
We use Logistic Regression from scikit-learn library \citep{pedregosa2011scikit} as a probing classifier. The classifier is trained over hidden representations\footnote{Morphosyntactic Features and Values: we take mean-pooled representations of the sub-word embeddings that correspond to a target word. Masked Token: we use embedding of a tokenizer-specific masked token. Perturbations: we use mean-pooled sentence representations.} produced by the encoders with the regularization parameter $L^2$  $\in [0.25, 0.5, 1, 2, 4]$ tuned on the validation set. The performance is evaluated by the ROC-AUC score.

\paragraph{Neuron Analysis}
\label{subsec:ind_neuron_analysis}
The neuron-level analysis allows retrieving a group of individual neurons that are most relevant to predict a linguistic property \citep{durrani2020analyzing}. Similarly, a linear classifier is trained over concatenated mean-pooled word/sentence representations using Elastic-net regularization \citep{zou2005regularization}, and with $L^1$ and $L^2$ $\lambda$'s $\in [0.1, \ldots, 1\emph{e}^{-5}]$ tuned on the validation set. The weights of the classifier are used to measure the relevance of each neuron.

\paragraph{Correlation Analysis} Canonical correlation analysis (\texttt{ckasim}) is a representation-level similarity measure that allows identifying pairs of layers of similar behavior \citep{wu2020similarity}. We use \texttt{[CLS]}-pooled intermediate representations to analyze the encoders. The measure is computed with the help of the publicly available code\footnote{\url{https://github.com/johnmwu/contextual-corr-analysis}}.

\subsection{Baselines}
\label{baselines}
We train Logistic Regression over the following count-based and distributive baseline features (see Section \ref{subsec:probe_methods}). We use N-gram range $\in [1,4]$ for each count-based baseline. Count-based features include \textbf{Char Number} (length of a word/sentence in characters), \textbf{TF-IDF over character N-grams}, \textbf{TF-IDF over BPE tokens} (BertTokenizer), and \textbf{TF-IDF over SentencePiece tokens} (XLMRobertaTokenizer). We use multilingual tokenizers by HuggingFace library to split words/sentences into the sub-word tokens. The distributive baseline is mean-pooled monolingual \textbf{fastText}\footnote{\url{https://fasttext.cc/docs/en/crawl-vectors.html}} word/sentence embeddings \cite{bojanowski2017enriching}.

\section{Results}
\label{results}

\begin{figure*}[ht!]
  \centering
  \includegraphics[width=.95\textwidth]{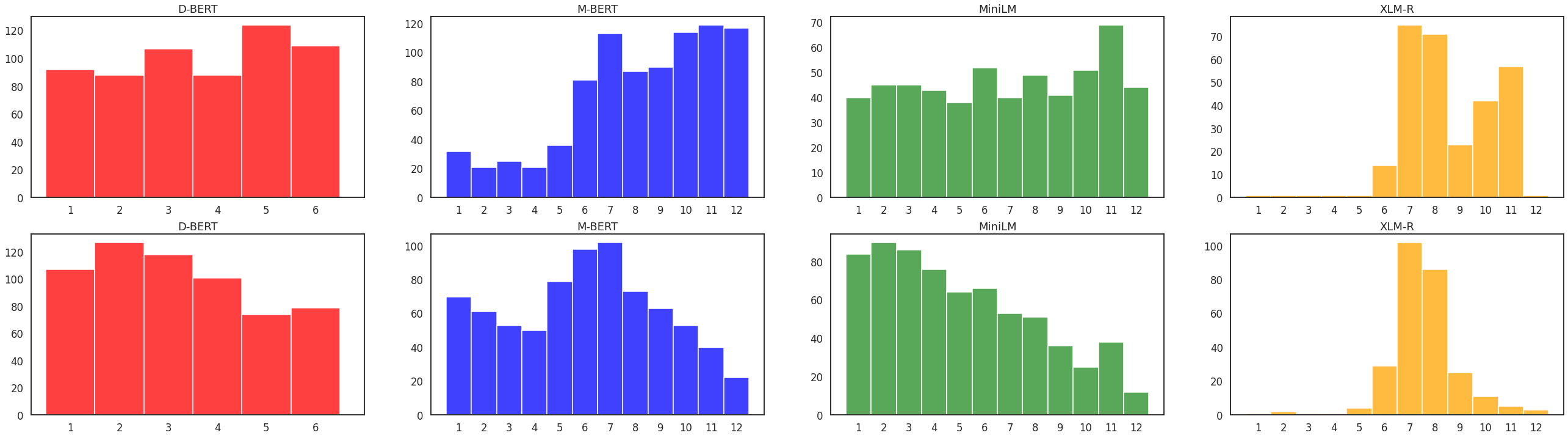}
  \caption{The distribution of top neurons over \textbf{Predicate Gender} perturbation task for each model. X-axis=Layer index number. Y-axis=Number of neurons. Top: pre-trained models. Bottom: fine-tuned models.}
  \label{ru_pred_gender_distribution}
\end{figure*}

\subsection{Morphosyntactic Features}
\label{result_features}
\paragraph{Probing Classifiers}

We learn the probing classifiers to estimate the model awareness of the morphosyntactic properties (see Section \ref{subsec:probe_methods}). The results demonstrate that pre-trained models perform slightly worse than their fine-tuned versions (2-4\%). We find that the awareness is distributed in a very similar manner despite the language differences, for the models of both instances (see Tables \ref{tab:feats_pretrained}--\ref{tab:feats_finetuned}, Appendix \ref{probing_perf}). Specifically, the performance on \textbf{Number} and \textbf{Gender} is reaching its plateau at the middle layers $[5-8]$ of 12-layer models, and at layer $[3]$ of \textbf{D-BERT}. The probing curves\footnote{We refer to a \emph{probing curve} as to a graphical representation of the probing classifier performance.} on \textbf{Case} are achieving their peak at the lower-to-middle layers $[4-5]$ and staying at the plateau towards the output layer. The only difference is observed on \textbf{Person} where the property is best inferred either across all layers (English, Russian) or at the lower-to-higher layers $[4-11]$ (French, German). The baseline features receive a strong performance, meaning that the occurrence of certain property may be inferred using the sub-word information (see Table \ref{tab:cat-baseline}, Appendix \ref{app:baseline_performance}).

\vspace{0.1cm}

\subsection{Masked Token} \label{result_masked_token}
\paragraph{Probing Classifiers}
The results of the probing classifier performance on {\bf Masked Token} tasks are presented in Tables~\ref{tab:mask-nottuned} (pre-trained models) and~\ref{tab:mask-tuned} (fine-tuned models) (see Appendix~\ref{probing_perf}). The task has appeared to be more challenging as opposed to \textbf{Morphosyntactic Features} (see Section \ref{result_features}). An interesting observation in this setting is that the performance of the models predominantly drops or becomes unstable after fine-tuning. For instance, \textbf{BERT} may lose almost $10\%$ in the tasks for Russian, and \textbf{D-BERT} may drop $5\%$ in the tasks for French. The probing curves tend to show rapid increases and decreases across the layers. An exception to this pattern is \textbf{XLM-R} which is less affected by fine-tuning and exhibits a more stable probing behavior. Nevertheless, the models demonstrate their capability to infer the properties from the context. \textbf{XLM-R} makes correct predictions in almost $70\%$ of cases, while the performance of \textbf{M-BERT} and \textbf{D-BERT} is slightly worse, and \textbf{MiniLM} may struggle the most. Figure~\ref{mask_ru_case} outlines the results on {\bf Case} task for Russian, best solved among the others. The middle-to-higher layers account for more correct predictions in the models of both instances. However, the higher layers $[10-12]$ of 12-layer models and layer $[6]$ of \textbf{D-BERT} may pertain to lower performance. A possible explanation is that the layers are affected by the objectives, i.e., Masked Language Modeling (pre-trained) or POS-tagging (fine-tuned). We find that the contextualized representations of a masked token produced by the final layers of pre-trained models may store the morphosyntactic properties. The probing curves demonstrate that the distribution of the properties may get affected by fine-tuning, or the knowledge can be partially lost, which is shown by the performance drops.

\subsection{Morphosyntactic Values}
\label{results_values}
\paragraph{Property-wise Neuron Analysis} We apply property-wise neuron analysis to investigate the top-neurons per each morphosyntactic property (see Section \ref{subsec:probe_methods}). We find that some models require a larger group of neurons to learn a morphosyntactic property, and the number of these neurons may get changed after fine-tuning. We provide the results for each language in Appendix \ref{app:individual_neuron_an}.
Figure \ref{french_prop_wise} illustrates the distributions for pre-trained and fine-tuned models for French. While after fine-tuning the number of neurons on \textbf{Person} (\textbf{M-BERT}, \textbf{D-BERT}) and \textbf{Number} (\textbf{XLM-R}) has increased, \textbf{Number} and \textbf{Gender} are now handled by fewer neurons of the distilled models (\textbf{D-BERT}, \textbf{MiniLM}). A similar behavior is observed for Russian and English. \textbf{Case} (Russian), \textbf{Gender} (Russian) and \textbf{Person} (English) require more neurons (\textbf{M-BERT}), or fewer neurons over \textbf{Person} (Russian) and \textbf{Number} (Russian, English) (\textbf{MiniLM}, \textbf{D-BERT}). Notably, the fine-tuning phase does not affect the neuron distributions for German.

\subsection{Perturbations}
\label{result_perturbations}
\paragraph{Probing Classifiers} The results of the probing classifier performance on \textbf{Perturbations} tasks are presented in Table \ref{err-nottuned} (pre-trained models), and Table \ref{err-tuned} (fine-tuned models) (see Appendix \ref{probing_perf}). We find that the models perform on par with one another in the majority of the tasks. Notably, \textbf{XLM-R} is generally the most sensitive to the perturbations in each language compared to the other models. We find that the syntactic perturbations (\textbf{Article Removal}, \textbf{Stopwords Removal}) are better solved than the inflectional ones. Similarly, the count-based baselines receive the best performance on the syntactic perturbations since the latter are obtained over a limited set of words (see Table \ref{tab:perturbation_baseline}, Appendix \ref{app:baseline_performance}). On the other hand, their performance is typically higher or close to random on the inflectional perturbations (see Table \ref{tab:perturbation_baseline}, Appendix \ref{app:baseline_performance}). We briefly describe the results in Appendix \ref{probing_perf} for the sake of space.

\paragraph{Layer-wise Neuron Analysis}
Individual neuron analysis helps to observe how top-neurons are spread across the entire model, and identify the relevance of each layer by the number of its top-neurons\footnote{We selected top-20\% neurons using the neuron ranking algorithm \citep{durrani-etal-2020-analyzing}.} (see Section \ref{subsec:ind_neuron_analysis}). Figure \ref{ru_pred_gender_distribution} demonstrates the results for \textbf{Predicate Gender} task in Russian. The sensitivity to the perturbation tends to be distributed across all layers of both pre-trained and fine-tuned models (\textbf{D-BERT}, \textbf{M-BERT}, \textbf{MiniLM}). The exception is provided by \textbf{XLM-R} which localizes the knowledge at the middle-to-higher layers $[6-11]$ (pre-trained), or in fewer layers but with larger groups of neurons $[6-9]$ (fine-tuned). The models of both instances store the sensitivity to the incorrect subject case form (\textbf{Subject Case}, Appendix \ref{app:individual_neuron_an}) at the middle-to-higher layers (\textbf{D-BERT}: $[3-5]$, \textbf{M-BERT}: $[6-11]$, \textbf{MiniLM}: $[4-8]$, \textbf{XLM-R}: $[5-12]$). Notably, the number of top-neurons in all models has decreased after the fine-tuning, and the information has been now more localized in two of them (\textbf{MiniLM}, \textbf{XLM-R}). A similar behavior of the models by language is observed on \textbf{Subject Number} (see Appendix \ref{app:individual_neuron_an}). The property is generally captured at the middle-to-higher layers of each pre-trained model for Russian, German and French (\textbf{D-BERT}: $[2, 3-6]$, \textbf{M-BERT}: $[6-12]$, \textbf{MiniLM}: $[5-12]$). The results are different for their fine-tuned versions, where the property gets more localized for Russian and German (\textbf{D-BERT}: $[3-5]$, \textbf{MiniLM}: $[5-7]$, \textbf{XLM-R}: $[6-9]$, \textbf{M-BERT}: $[6-11]$), or captured by fewer neurons at the same layers for French. In contrast, the property is predominantly distributed across all layers of both pre-trained and fine-tuned models for English.

\begin{figure}[ht!]
\centering
\includegraphics[width=0.45\textwidth]{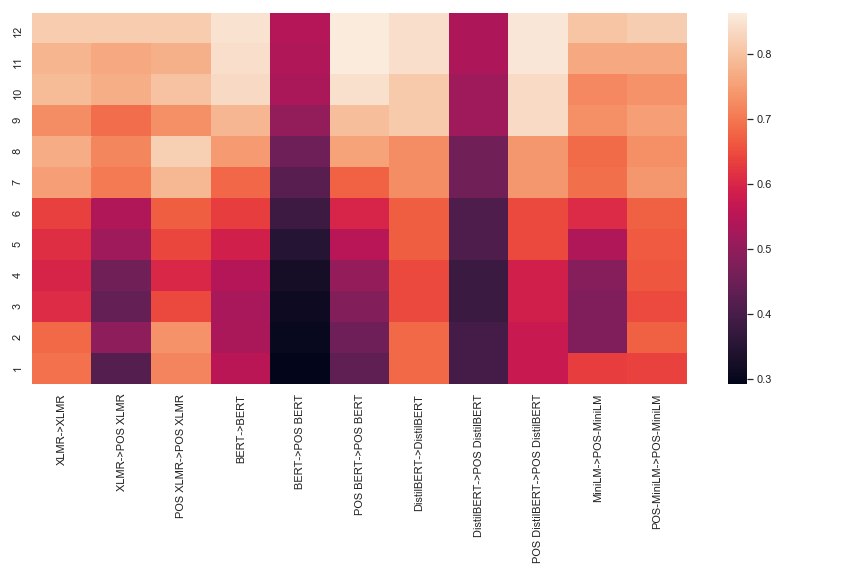}
\caption{\texttt{ckasim} results on \textbf{Stopwords Removal} task in German. X-axis=Model instance combinations. Y-axis=Layer index number (left), \texttt{ckasim} score (right).}
\label{cca_de_stopwords}
\end{figure}

\paragraph{Correlation Analysis} To analyze the encoders with \texttt{ckasim}, we take \texttt{[CLS]}-pooled representations of the original sentence (without the perturbation) and its perturbed version. The similarity measure is computed on the resulted pairs of representations. For each model \textbf{M} we explore three settings by combining different model instances (see Section \ref{models}): (i) (\textit{pre-trained} \textbf{M}, \textit{pre-trained} \textbf{M}), (ii) (\textit{pre-trained} \textbf{M}, \textit{fine-tuned} \textbf{M}), (iii) (\textit{fine-tuned} \textbf{M}, \textit{fine-tuned} \textbf{M}). Figure \ref{cca_de_stopwords} shows the most typical pattern achieved in the tasks. The biggest difference is observed over the combination (ii), where the perturbations are best captured at the lower-to-middle layers $[1-6]$ (\textbf{XLM-R}, \textbf{MiniLM}), or across all the layers (\textbf{M-BERT}, \textbf{DistilBERT}). The middle-to-higher layers $[7-12]$ tend to become more similar over combinations (i, iii) which may mean that they are able to restore the semantics of the perturbed sentences, being more robust to the perturbations as opposed to the lower ones.

\section{Discussion}
\label{discussion}
\paragraph{Morphosyntactic content across languages}
The probing curves under layer-wise probing demonstrate that the multilingual transformers learn the morphosyntactic content in a greatly similar manner despite the language differences (see Section \ref{result_features}). The properties are predominantly distributed across the middle-to-higher layers $[5-12]$ for each language. In contrast, \textbf{Masked Token} tasks represent a challenge for the models causing rapid increases and decreases in the performance across the layers (see Section \ref{result_masked_token}). The overall pattern for each language is that a masked token's properties are best inferred at the middle-to-higher layers. A possible reason for this is that the task requires incorporating syntactic and semantic information from the context since the target word remains unseen. The models demonstrate their sensitivity to \textbf{Perturbations} (see Section \ref{result_perturbations}). While the syntactic perturbations are predominantly captured at the lower-to-middle layers $[3-8]$, the inflectional ones are stored at the middle-to-higher layers $[5-12]$. In contrast to other languages, the perturbation properties for English may be distributed across all layers of the models. The results are supported by the individual neuron analysis, an example of which is provided in Appendix \ref{app:individual_neuron_an}.

\paragraph{Same properties require different number of neurons} Property-wise neuron analysis shows that \textbf{Person} and \textbf{Case} are learned using more neurons as compared to \textbf{Number} and \textbf{Gender} across the languages. Notably, the number of neurons required to learn a property may depend on the language. For example, \textbf{D-BERT} requires about 1000 neurons to learn \textbf{Case} in German and less than 1500 neurons to learn the property in Russian.

\paragraph{Are students good learners?}
A common method to compare pre-trained models and their distilled versions is based upon their performance on downstream tasks \citep{tsai2019small}, or NLU benchmarks \citep{wang2018glue,wang2019superglue}. Still, little is investigated on what language properties are preserved after the knowledge distillation. We find that \textbf{D-BERT} and \textbf{MiniLM} mimic the behavior of their teachers under layer-wise probing (see Section \ref{result_features}), or display a similar perturbation sensitivity under \texttt{ckasim} (see Section \ref{result_perturbations}). However, \textbf{MiniLM} tends to exhibit an uncertain behavior as opposed to their teacher (see Sections \ref{result_masked_token}, \ref{result_perturbations}).

\paragraph{Effect of fine-tuning}
The results show that the effect of fine-tuning for POS-tagging varies within a certain group of tasks. First, fine-tuned models may receive a better probing performance by 2-4\% on \textbf{Morphosyntactic Features} tasks (see Section \ref{result_features}). Second, fine-tuning affects the way the properties are distributed or causes significant performance drops on \textbf{Masked Token} tasks, specifically at the higher layers (see Section \ref{result_masked_token}). The impact on the property distribution is also demonstrated on \textbf{Perturbations} tasks under neuron-level probe (see Section \ref{result_perturbations}). Besides, the analysis of top-neurons allows concluding that fine-tuning may affect localization (\textbf{MiniLM}, \textbf{XLM-R}) which is in line with \citep{wu2020similarity}. Finally, a number of neurons required to predict a property may increase (e.g., Russian: \textbf{Case}; French: \textbf{Person}), decrease (e.g., English: \textbf{Number}) or remain unchanged (German). We suggest that an interesting line for future work is to analyze the correlation between the number of neurons and the probe performance after fine-tuning. For instance, the results on \textbf{Perturbation} tasks indicate that some models may receive a better probing performance with fewer (\textbf{XLM-R}) or more neurons (\textbf{D-BERT}, \textbf{M-BERT}) (see Section \ref{result_perturbations}). An exploration of fine-tuning for morphosyntactic analysis, specifically over UniMorph \citep{kirov2018unimorph} may be a fruitful avenue for future work.

\paragraph{Distribution of knowledge may depend on language morphology} The analysis of the models under layer-wise and neuron-wise probing suggests that the behavior may depend on how morphologically rich a language is (see Sections \ref{result_features}, \ref{result_perturbations}). The knowledge for English tends to be distributed across all layers of the models in contrast to the more morphologically rich languages that capture the properties at the middle-to-higher layers. The finding is in line with a few recent studies \citep{edmiston2020systematic,durrani-etal-2020-analyzing,elazar2020bert} which contradict the common understanding that morphology is stored at the lower layers \citep{tenney-etal-2019-bert,rogers-etal-2020-primer}. We also find that the distribution of the properties varies based on the complexity of a probing task (see Sections \ref{result_features}, \ref{result_masked_token}). An exciting direction for future work is to test this hypothesis on a more diverse set of morphologically contrasting languages. Besides, perturbing one aspect of a sentence can cause ambiguity elsewhere which is an interesting line for future exploration of the interdependence of the perturbations.

\section{Conclusion}
\label{conclusion}
This paper proposes \textbf{Morph Call}, a suite of 46 probing tasks in four Indo-European languages that differ significantly in their richness of morphology: Russian, French, English, and German. The suite includes a new type of probing task based on the detection of syntactic and inflectional sentence perturbations. 
We apply a combination of three introspection methods based on neuron-, layer- and representation-level analysis to probe five multilingual transformer models, including their less explored distilled versions. The analysis of transformers' understudied aspect contradicts the common findings on how morphology is represented in the models. We find that the knowledge for English is predominantly distributed across all layers of the models in contrast to more morphologically rich languages (German, Russian, French), which house the properties at the middle-to-higher layers. The models demonstrate their sensitivity to the perturbations, and \textbf{XLM-R} tends to be the most robust among the others. We observe that distilled models inherit their teachers' knowledge, showing a comparative performance and exhibiting similar property distribution on several probing tasks. Another finding is that fine-tuning for POS-tagging can affect the model knowledge in various manners, ranging from improving and decreasing the probing classifier performance to changing the information's localization. We believe there is still room for exploring the models' morphosyntactic content and the effect of fine-tuning, specifically across a more diverse set of languages and types of model architectures.

\section*{Acknowledgements}
We thank our reviewers for their insightful comments and suggestions. Ekaterina Artemova is supported by the framework of the HSE University Basic Research Program.

\bibliography{anthology,custom}
\bibliographystyle{acl_natbib}

\appendix
\newpage
\onecolumn
\setcounter{table}{0}
\setcounter{figure}{0}

\section*{Appendix}
\section{Description of Treebanks}
\label{app:Treebank_desc}
Below is a list of the UD Treebanks used in the experiments:

\begin{itemize}
    \item \textbf{Russian}: GramEval2020 Treebanks, GSD Russian Treebank, Russian-PUD, and SynTagRus Treebank.
    \item \textbf{English}: EWT Treebank, GUM
Treebank, the English portion of ParTUT, English-PUD, and English-Pronouns Treebank.
    \item \textbf{French}: French Question Bank, GSD French Treebank, the French portion of ParTUT, French-PUD, Sequoia and  French Spoken Treebank, adapted from the Rhapsoide prosodic-syntactic Treebank.
    \item \textbf{German}: GSD German Treebank, HDT-UD Treebank, German-PUD and LIT German Treebank.
\end{itemize}

\newpage

\section{Dataset Statistics}
\label{app:data_stata}
Tables \ref{tab:morph_feat_mask_stata} -- \ref{tab:perturbation_stata} provide a brief statistics on the partition sizes for each probing task. 

\begin{table*}[h]
\centering
\begin{tabular}{c|c|c|c|c|c}
\toprule
\textbf{Probing Task} & \textbf{Language} & \textbf{Train} & \textbf{Dev} & \textbf{Test} &  \textbf{Overall}\\
\midrule
\textbf{Number} & 
    \begin{tabular}{@{}c@{}c@{}c@{}c}\textbf{Ru} \\ \textbf{En} \\ \textbf{De} \\ \textbf{Fr}\end{tabular}
    &
    \begin{tabular}{@{}c@{}c@{}c@{}c}
        174 720 \\ 51 465 \\ 533 898 \\ 74 450
    \end{tabular}
    &
    \begin{tabular}{@{}c@{}c@{}c@{}c}
         21 937 \\ 6492 \\ 66 984 \\ 9385
    \end{tabular}
    &
    \begin{tabular}{@{}c@{}c@{}c@{}c}
         21 379 \\ 6374 \\ 66 984 \\ 9191 
    \end{tabular}
    &
    \begin{tabular}{@{}c@{}c@{}c@{}c}
         218 036 \\ 64 331 \\ 668 271 \\ 93 026 
    \end{tabular} \\ \midrule

\textbf{Case} & 
    \begin{tabular}{@{}c@{}c}\textbf{Ru} \\ \textbf{De} \end{tabular}
    &
    \begin{tabular}{@{}c@{}c}
        174 884 \\ 436 303
    \end{tabular}
    &
    \begin{tabular}{@{}c@{}c}
         21 768 \\ 54 692
    \end{tabular}
    &
    \begin{tabular}{@{}c@{}c}
         21 974 \\ 53 932
    \end{tabular}
    &
    \begin{tabular}{@{}c@{}c}
         218 626 \\ 544 927
    \end{tabular} \\ \midrule

\textbf{Person} & 
    \begin{tabular}{@{}c@{}c@{}c@{}c}\textbf{Ru} \\ \textbf{En} \\ \textbf{De} \\ \textbf{Fr}\end{tabular}
    &
    \begin{tabular}{@{}c@{}c@{}c@{}c}
        162 345 \\ 47 001 \\ 471 132 \\ 71 394
    \end{tabular}
    &
    \begin{tabular}{@{}c@{}c@{}c@{}c}
         20 313 \\ 5945 \\ 58 847 \\ 8853
    \end{tabular}
    &
    \begin{tabular}{@{}c@{}c@{}c@{}c}
         20 319 \\ 5735 \\ 58 438 \\ 8992 
    \end{tabular}
    &
    \begin{tabular}{@{}c@{}c@{}c@{}c}
         202 977 \\ 58 681 \\ 588 417 \\ 89 239 
    \end{tabular} \\ \midrule

\textbf{Gender} & 
    \begin{tabular}{@{}c@{}c@{}c}\textbf{Ru} \\ \textbf{De} \\ \textbf{Fr}\end{tabular}
    &
    \begin{tabular}{@{}c@{}c@{}c}
        165 934 \\ 500 628 \\ 69 901
    \end{tabular}
    &
    \begin{tabular}{@{}c@{}c@{}c}
         20 462 \\ 62 163 \\ 8840
    \end{tabular}
    &
    \begin{tabular}{@{}c@{}c@{}c}
         20 982 \\ 62 612 \\ 8559
    \end{tabular}
    &
    \begin{tabular}{@{}c@{}c@{}c}
         207 378 \\ 625 403 \\ 87 300
    \end{tabular} \\ \bottomrule

\end{tabular}
\caption{Number of samples for each \textbf{Morphosyntactic Features} and \textbf{Masked Token} task. Languages: \textbf{Ru}=Russian, \textbf{En}=English, \textbf{De}=German, \textbf{Fr}=French.}
\label{tab:morph_feat_mask_stata}
\end{table*}

\begin{table*}[h]
\centering
\begin{tabular}{c|c|c|c|c|c}
\toprule
\textbf{Probing Task} & \textbf{Language} & \textbf{Train} & \textbf{Dev} & \textbf{Test} &  \textbf{Overall}\\
\midrule
\textbf{Number} & 
    \begin{tabular}{@{}c@{}c@{}c@{}c}\textbf{Ru} \\ \textbf{En} \\ \textbf{De} \\ \textbf{Fr}\end{tabular}
    &
    \begin{tabular}{@{}c@{}c@{}c@{}c}
        100 738 \\ 21 568 \\ 339 744 \\ 33 339
    \end{tabular}
    &
    \begin{tabular}{@{}c@{}c@{}c@{}c}
         12 592 \\ 2696 \\ 42 468 \\ 4167
    \end{tabular}
    &
    \begin{tabular}{@{}c@{}c@{}c@{}c}
         12 593 \\ 2696 \\ 42 468 \\ 4168 
    \end{tabular}
    &
    \begin{tabular}{@{}c@{}c@{}c@{}c}
         125 923 \\ 26 960 \\ 424 680 \\ 41 674
    \end{tabular} \\ \midrule

\textbf{Case} & 
    \begin{tabular}{@{}c@{}c}\textbf{Ru} \\ \textbf{De} \end{tabular}
    &
    \begin{tabular}{@{}c@{}c}
        92 320 \\ 252 182
    \end{tabular}
    &
    \begin{tabular}{@{}c@{}c}
         11 540 \\ 31 523
    \end{tabular}
    &
    \begin{tabular}{@{}c@{}c}
        11 540 \\ 31 523
    \end{tabular}
    &
    \begin{tabular}{@{}c@{}c}
         115 400 \\ 315 228
    \end{tabular} \\ \midrule

\textbf{Person} & 
    \begin{tabular}{@{}c@{}c@{}c@{}c}\textbf{Ru} \\ \textbf{En} \\ \textbf{De} \\ \textbf{Fr}\end{tabular}
    &
    \begin{tabular}{@{}c@{}c@{}c@{}c}
        15 748 \\ 7255 \\ 184 788 \\ 6364
    \end{tabular}
    &
    \begin{tabular}{@{}c@{}c@{}c@{}c}
         11 540 \\ 907 \\ 23 099 \\ 796
    \end{tabular}
    &
    \begin{tabular}{@{}c@{}c@{}c@{}c}
         11 540 \\ 907 \\ 23 099 \\ 796
    \end{tabular}
    &
    \begin{tabular}{@{}c@{}c@{}c@{}c}
         19 685 \\ 9069 \\ 230 986 \\ 7956 
    \end{tabular} \\ \midrule

\textbf{Gender} & 
    \begin{tabular}{@{}c@{}c@{}c}\textbf{Ru} \\ \textbf{De} \\ \textbf{Fr}\end{tabular}
    &
    \begin{tabular}{@{}c@{}c@{}c}
        76 158 \\ 252 182 \\ 23 660
    \end{tabular}
    &
    \begin{tabular}{@{}c@{}c@{}c}
         9520 \\ 31 523 \\ 2957
    \end{tabular}
    &
    \begin{tabular}{@{}c@{}c@{}c}
         9520 \\ 31 523 \\ 2958
    \end{tabular}
    &
    \begin{tabular}{@{}c@{}c@{}c}
         95 198 \\ 315 228 \\ 29 575
    \end{tabular} \\ \bottomrule

\end{tabular}
\caption{Number of samples for each \textbf{Morphosyntactic Values} task. Languages: \textbf{Ru}=Russian, \textbf{En}=English, \textbf{De}=German, \textbf{Fr}=French.}
\label{tab:morph_values_stata}
\end{table*}

\begin{table*}[h]
\centering
\begin{tabular}{c|c|c|c|c|c}
\toprule
\textbf{Probing Task} & \textbf{Language} & \textbf{Train} & \textbf{Dev} & \textbf{Test} &  \textbf{Overall}\\
\midrule
\textbf{Stop-words Removal} & 
    \begin{tabular}{@{}c@{}c@{}c@{}c}\textbf{Ru} \\ \textbf{En} \\ \textbf{De} \\ \textbf{Fr}\end{tabular}
    &
    \begin{tabular}{@{}c@{}c@{}c@{}c}
        38 838 \\ 12 627 \\ 121 272 \\ 13 959
    \end{tabular}
    &
    \begin{tabular}{@{}c@{}c@{}c@{}c}
         4855 \\ 1578 \\ 15 159 \\ 1745
    \end{tabular}
    &
    \begin{tabular}{@{}c@{}c@{}c@{}c}
         4855 \\ 1578 \\ 15 159 \\ 1745 
    \end{tabular}
    &
    \begin{tabular}{@{}c@{}c@{}c@{}c}
         48 548 \\ 15 784 \\ 151 590 \\ 17 449 
    \end{tabular} \\ \midrule
 
\textbf{Article Removal} & 
    \begin{tabular}{@{}c@{}c@{}c}\textbf{En} \\ \textbf{De} \\ \textbf{Fr}\end{tabular}
    &
    \begin{tabular}{@{}c@{}c@{}c}
        7770 \\ 99 669 \\ 10 083 
    \end{tabular}
    &
    \begin{tabular}{@{}c@{}c@{}c}
         971 \\ 12459 \\ 1253
    \end{tabular}
    &
    \begin{tabular}{@{}c@{}c@{}c}
         972 \\ 12459 \\ 1276
    \end{tabular}
    &
    \begin{tabular}{@{}c@{}c@{}c}
         15 784 \\ 124 587 \\ 12 612
    \end{tabular} \\ \midrule
    
\textbf{Subject Number} & 
    \begin{tabular}{@{}c@{}c@{}c@{}c}\textbf{Ru} \\ \textbf{En} \\ \textbf{De} \\ \textbf{Fr}\end{tabular}
    &
    \begin{tabular}{@{}c@{}c@{}c@{}c}
        9293 \\ 471 \\ 5 709 \\ 1219
    \end{tabular}
    &
    \begin{tabular}{@{}c@{}c@{}c@{}c}
         1164 \\ 58 \\ 1007 \\ 151
    \end{tabular}
    &
    \begin{tabular}{@{}c@{}c@{}c@{}c}
         1165 \\ 60 \\ 1009 \\ 153 
    \end{tabular}
    &
    \begin{tabular}{@{}c@{}c@{}c@{}c}
         11 622 \\ 589 \\ 6005 \\ 1523
    \end{tabular} \\ \midrule
    
\textbf{Subject Case} & 
    \begin{tabular}{@{}c}\textbf{Ru}\end{tabular}
    &
    \begin{tabular}{@{}c}
        18 897
    \end{tabular}
    &
    \begin{tabular}{@{}c}
        2344
    \end{tabular}
    &
    \begin{tabular}{@{}c}
        2346
    \end{tabular}
    &
    \begin{tabular}{@{}c}
        23 587
    \end{tabular} \\ \midrule
    
\textbf{Predicate Number} & 
    \begin{tabular}{@{}c@{}c@{}c@{}c}\textbf{Ru} \\ \textbf{En} \\ \textbf{De} \\ \textbf{Fr}\end{tabular}
    &
    \begin{tabular}{@{}c@{}c@{}c@{}c}
        7160 \\ 1115 \\ 26 415 \\ 2822
    \end{tabular}
    &
    \begin{tabular}{@{}c@{}c@{}c@{}c}
         897 \\ 140 \\ 4374 \\ 353
    \end{tabular}
    &
    \begin{tabular}{@{}c@{}c@{}c@{}c}
         897 \\ 142 \\ 4375 \\ 356 
    \end{tabular}
    &
    \begin{tabular}{@{}c@{}c@{}c@{}c}
         8 954 \\ 1397 \\ 35 164 \\ 3531
    \end{tabular} \\ \midrule

\textbf{Predicate Person} & 
    \begin{tabular}{@{}c}\textbf{Ru}\end{tabular}
    &
    \begin{tabular}{@{}c}
        5240
    \end{tabular}
    &
    \begin{tabular}{@{}c}
        644
    \end{tabular}
    &
    \begin{tabular}{@{}c}
        646
    \end{tabular}
    &
    \begin{tabular}{@{}c}
        6530
    \end{tabular} \\ \midrule

\textbf{Predicate Gender} & 
    \begin{tabular}{@{}c}\textbf{Ru}\end{tabular}
    &
    \begin{tabular}{@{}c}
        4414
    \end{tabular}
    &
    \begin{tabular}{@{}c}
        550
    \end{tabular}
    &
    \begin{tabular}{@{}c}
        553
    \end{tabular}
    &
    \begin{tabular}{@{}c}
        5517
    \end{tabular} \\ \midrule

\textbf{Deixis Word Number} & 
    \begin{tabular}{@{}c@{}c}\textbf{En} \\ \textbf{De} \end{tabular}
    &
    \begin{tabular}{@{}c@{}c}
        1130 \\ 4804
    \end{tabular}
    &
    \begin{tabular}{@{}c@{}c}
         141 \\ 600
    \end{tabular}
    &
    \begin{tabular}{@{}c@{}c}
         142 \\ 601
    \end{tabular}
    &
    \begin{tabular}{@{}c@{}c}
         1413 \\ 6005
    \end{tabular} \\ \bottomrule

\end{tabular}
\caption{Number of samples for each \textbf{Perturbation} task. Languages: \textbf{Ru}=Russian, \textbf{En}=English, \textbf{De}=German, \textbf{Fr}=French.}
\label{tab:perturbation_stata}
\end{table*}

\clearpage

\section{Baseline Performance}
\label{app:baseline_performance}
Table \ref{tab:cat-baseline} summarizes the results of the baseline models for \textbf{Morphosyntactic Features} tasks. Table \ref{tab:perturbation_baseline} presents the performance of the baseline models for \textbf{Perturbations} tasks.

\begin{table*}[h]
\centering
\resizebox{\textwidth}{!}{
\begin{tabular}{c|c|c|c|c|c|c}
\toprule
\textbf{Probing Task} & \textbf{Lang} & \textbf{Char Num} & \textbf{TF-IDF Char} & \textbf{TF-IDF BPE} &  \textbf{TF-IDF SP} & \textbf{fT}\\
\midrule
\textbf{Number} & 
    \begin{tabular}{@{}c@{}c@{}c@{}c@{}c}
    \textbf{Ru} \\ \textbf{En} \\ \textbf{De} \\ \textbf{Fr}\end{tabular}
    &
    \begin{tabular}{@{}c@{}c@{}c@{}c@{}c}
        0.78 \\ 0.63 \\ 0.57 \\ 0.52
    \end{tabular}
    &
    \begin{tabular}{@{}c@{}c@{}c@{}c@{}c}
         \textbf{0.97} \\ \textbf{0.95} \\ \textbf{0.95} \\ \textbf{0.91}
    \end{tabular}
    &
    \begin{tabular}{@{}c@{}c@{}c@{}c@{}c}
         0.96 \\ 0.94 \\ \textbf{0.95} \\ \textbf{0.91} 
    \end{tabular}
    &
    \begin{tabular}{@{}c@{}c@{}c@{}c@{}c}
         0.96 \\ \textbf{0.95} \\ 0.95 \\ \textbf{0.91}
    \end{tabular}
    &
    \begin{tabular}{@{}c@{}c@{}c@{}c@{}c}
         0.94 \\ 0.93 \\ 0.89 \\ 0.87
    \end{tabular}
    \\ \midrule

\textbf{Case} & 
    \begin{tabular}{@{}c@{}c}\textbf{Ru} \\ \textbf{De} \end{tabular}
    &
    \begin{tabular}{@{}c@{}c}
        0.69 \\ 0.64
    \end{tabular}
    &
    \begin{tabular}{@{}c@{}c}
         \textbf{0.97} \\ 0.92
    \end{tabular}
    &
    \begin{tabular}{@{}c@{}c}
         0.96 \\ \textbf{0.93}
    \end{tabular}
    &
    \begin{tabular}{@{}c@{}c}
         0.96 \\ 0.92
    \end{tabular}
    &
    \begin{tabular}{@{}c@{}c}
         0.90 \\ 0.88
    \end{tabular}
    \\ \midrule

\textbf{Person} & 
    \begin{tabular}{@{}c@{}c@{}c@{}c@{}c}\textbf{Ru} \\ \textbf{En} \\ \textbf{De} \\ \textbf{Fr}\end{tabular}
    &
    \begin{tabular}{@{}c@{}c@{}c@{}c@{}c}
        0.60 \\ 0.62 \\ 0.66 \\ 0.54
    \end{tabular}
    &
    \begin{tabular}{@{}c@{}c@{}c@{}c@{}c}
         \textbf{0.98} \\ 0.97 \\ \textbf{0.93} \\ \textbf{0.93}
    \end{tabular}
    &
    \begin{tabular}{@{}c@{}c@{}c@{}c@{}c}
         \textbf{0.98} \\ 0.97 \\ \textbf{0.93} \\ 0.92
    \end{tabular}
    &
    \begin{tabular}{@{}c@{}c@{}c@{}c@{}c}
         \textbf{0.98} \\ 0.97 \\ \textbf{0.93} \\ 0.92 
    \end{tabular}
    &
    \begin{tabular}{@{}c@{}c@{}c@{}c@{}c}
         0.93 \\ \textbf{0.98} \\ 0.91 \\ 0.88 
    \end{tabular}
    \\ \midrule

\textbf{Gender} & 
    \begin{tabular}{@{}c@{}c@{}c}\textbf{Ru} \\ \textbf{De} \\ \textbf{Fr}\end{tabular}
    &
    \begin{tabular}{@{}c@{}c@{}c}
        0.73 \\ 0.47 \\ 0.54
    \end{tabular}
    &
    \begin{tabular}{@{}c@{}c@{}c}
         \textbf{0.96} \\ \textbf{0.86} \\ \textbf{0.88}
    \end{tabular}
    &
    \begin{tabular}{@{}c@{}c@{}c}
         0.95 \\ \textbf{0.86} \\ \textbf{0.88}
    \end{tabular}
    &
    \begin{tabular}{@{}c@{}c@{}c}
         \textbf{0.96} \\ \textbf{0.86} \\ 0.87
    \end{tabular}
    &
    \begin{tabular}{@{}c@{}c@{}c}
         0.89 \\ 0.81 \\ 0.84
    \end{tabular}
    \\ \bottomrule

\end{tabular}}
\caption{Baseline results on \textbf{Morphosyntactic Features} tasks. \textbf{SP} refers to SentencePiece, and \textbf{fT} corresponds to fastText. Languages: \textbf{Ru}=Russian, \textbf{En}=English, \textbf{De}=German, \textbf{Fr}=French.}
\label{tab:cat-baseline}
\end{table*}

\begin{table*}[h]
\centering
\resizebox{\textwidth}{!}{
\begin{tabular}{c|c|c|c|c|c|c}
\toprule
\textbf{Probing Task} & \textbf{Lang} & \textbf{Char Num} & \textbf{TF-IDF Char} & \textbf{TF-IDF BPE} &  \textbf{TF-IDF SP} & \textbf{fT} \\
\midrule
\textbf{Stop-words Removal} & 
    \begin{tabular}{@{}c@{}c@{}c@{}c}\textbf{Ru} \\ \textbf{En} \\ \textbf{De} \\ \textbf{Fr}\end{tabular}
    &
    \begin{tabular}{@{}c@{}c@{}c@{}c}
        0.57 \\ 0.64 \\ 0.63 \\ 0.60
    \end{tabular}
    &
    \begin{tabular}{@{}c@{}c@{}c@{}c}
         \textbf{0.96} \\ 0.97 \\ \textbf{0.99} \\ \textbf{0.98}
    \end{tabular}
    &
    \begin{tabular}{@{}c@{}c@{}c@{}c}
         0.92 \\ \textbf{0.98} \\ \textbf{0.99} \\ \textbf{0.98}
    \end{tabular}
    &
    \begin{tabular}{@{}c@{}c@{}c@{}c}
         0.92 \\ 0.97 \\ \textbf{0.99} \\ \textbf{0.98}
    \end{tabular}
    &
    \begin{tabular}{@{}c@{}c@{}c@{}c}
         0.93 \\ 0.96 \\ 0.97 \\ 0.96
    \end{tabular}
    \\ \midrule
 
\textbf{Article Removal} & 
    \begin{tabular}{@{}c@{}c@{}c}\textbf{En} \\ \textbf{De} \\ \textbf{Fr}\end{tabular}
    &
    \begin{tabular}{@{}c@{}c@{}c}
        0.52 \\ 0.55 \\ 0.56
    \end{tabular}
    &
    \begin{tabular}{@{}c@{}c@{}c}
         0.98 \\ \textbf{0.97} \\ 0.95
    \end{tabular}
    &
    \begin{tabular}{@{}c@{}c@{}c}
         \textbf{0.99} \\ \textbf{0.97} \\ \textbf{0.97}
    \end{tabular}
    &
    \begin{tabular}{@{}c@{}c@{}c}
         0.98 \\ \textbf{0.97} \\ 0.96
    \end{tabular}
    &
    \begin{tabular}{@{}c@{}c@{}c}
         0.84 \\ 0.87 \\ 0.87
    \end{tabular}
    \\ \midrule
    
\textbf{Subject Number} & 
    \begin{tabular}{@{}c@{}c@{}c@{}c}\textbf{Ru} \\ \textbf{En} \\ \textbf{De} \\ \textbf{Fr}\end{tabular}
    &
    \begin{tabular}{@{}c@{}c@{}c@{}c}
        0.50 \\ \textbf{0.43} \\ 0.5 \\ 0.44
    \end{tabular}
    &
    \begin{tabular}{@{}c@{}c@{}c@{}c}
         0.54 \\ 0.35 \\ 0.48 \\ \textbf{0.60}
    \end{tabular}
    &
    \begin{tabular}{@{}c@{}c@{}c@{}c}
         \textbf{0.55} \\ 0.37 \\ 0.46 \\ 0.50
    \end{tabular}
    &
    \begin{tabular}{@{}c@{}c@{}c@{}c}
         0.54 \\ \textbf{0.43} \\ 0.48 \\ 0.55
    \end{tabular} 
    &
    \begin{tabular}{@{}c@{}c@{}c@{}c}
         0.53 \\ 0.40 \\ \textbf{0.57} \\ 0.55
    \end{tabular}
    \\ \midrule
    
\textbf{Subject Case} & 
    \begin{tabular}{@{}c}\textbf{Ru}\end{tabular}
    &
    \begin{tabular}{@{}c}
        0.51
    \end{tabular}
    &
    \begin{tabular}{@{}c}
        \textbf{0.67}
    \end{tabular}
    &
    \begin{tabular}{@{}c}
        0.62
    \end{tabular}
    &
    \begin{tabular}{@{}c}
        0.62
    \end{tabular}
    &
    \begin{tabular}{@{}c}
        0.60
    \end{tabular}
    \\ \midrule
    
\textbf{Predicate Number} & 
    \begin{tabular}{@{}c@{}c@{}c@{}c}\textbf{Ru} \\ \textbf{En} \\ \textbf{De} \\ \textbf{Fr}\end{tabular}
    &
    \begin{tabular}{@{}c@{}c@{}c@{}c}
        0.49 \\ \textbf{0.52} \\ 0.50 \\ 0.49
    \end{tabular}
    &
    \begin{tabular}{@{}c@{}c@{}c@{}c}
         \textbf{0.64} \\ 0.49 \\ 0.60 \\ 0.64
    \end{tabular}
    &
    \begin{tabular}{@{}c@{}c@{}c@{}c}
         0.48 \\ 0.45 \\ 0.39 \\ 0.47
    \end{tabular}
    &
    \begin{tabular}{@{}c@{}c@{}c@{}c}
         0.50 \\ 0.47 \\ 0.38 \\ 0.49
    \end{tabular}
    &
    \begin{tabular}{@{}c@{}c@{}c@{}c}
         0.52 \\ 0.48 \\ \textbf{0.68} \\ \textbf{0.68}
    \end{tabular}
    \\ \midrule

\textbf{Predicate Person} & 
    \begin{tabular}{@{}c}\textbf{Ru}\end{tabular}
    &
    \begin{tabular}{@{}c}
        0.50
    \end{tabular}
    &
    \begin{tabular}{@{}c}
        \textbf{0.81}
    \end{tabular}
    &
    \begin{tabular}{@{}c}
        0.78
    \end{tabular}
    &
    \begin{tabular}{@{}c}
        0.74
    \end{tabular}
    &
    \begin{tabular}{@{}c}
        0.62
    \end{tabular}
    \\ \midrule

\textbf{Predicate Gender} & 
    \begin{tabular}{@{}c}\textbf{Ru}\end{tabular}
    &
    \begin{tabular}{@{}c}
        0.50
    \end{tabular}
    &
    \begin{tabular}{@{}c}
        \textbf{0.62}
    \end{tabular}
    &
    \begin{tabular}{@{}c}
        0.57
    \end{tabular}
    &
    \begin{tabular}{@{}c}
        0.58
    \end{tabular}
    &
    \begin{tabular}{@{}c}
        0.51
    \end{tabular}
    \\ \midrule

\textbf{Deixis Word Number} & 
    \begin{tabular}{@{}c@{}c}\textbf{En} \\ \textbf{De} \end{tabular}
    &
    \begin{tabular}{@{}c@{}c}
        0.48 \\ 0.49
    \end{tabular}
    &
    \begin{tabular}{@{}c@{}c}
         0.71 \\ 0.68
    \end{tabular}
    &
    \begin{tabular}{@{}c@{}c}
         \textbf{0.77} \\ 0.71
    \end{tabular}
    &
    \begin{tabular}{@{}c@{}c}
         0.75 \\ \textbf{0.72}
    \end{tabular}
    &
    \begin{tabular}{@{}c@{}c}
         0.70 \\ 0.62
    \end{tabular}
    \\ \bottomrule

\end{tabular}}
\caption{Baseline results on \textbf{Perturbation} tasks. \textbf{SP} refers to SentencePiece, and \textbf{fT} corresponds to fastText. Languages: \textbf{Ru}=Russian, \textbf{En}=English, \textbf{De}=German, \textbf{Fr}=French.}
\label{tab:perturbation_baseline}
\end{table*}

\clearpage

\section{Probing Classifiers}

\label{probing_perf} 
\begin{figure*}[h]
  \centering
  \includegraphics[width=\textwidth]{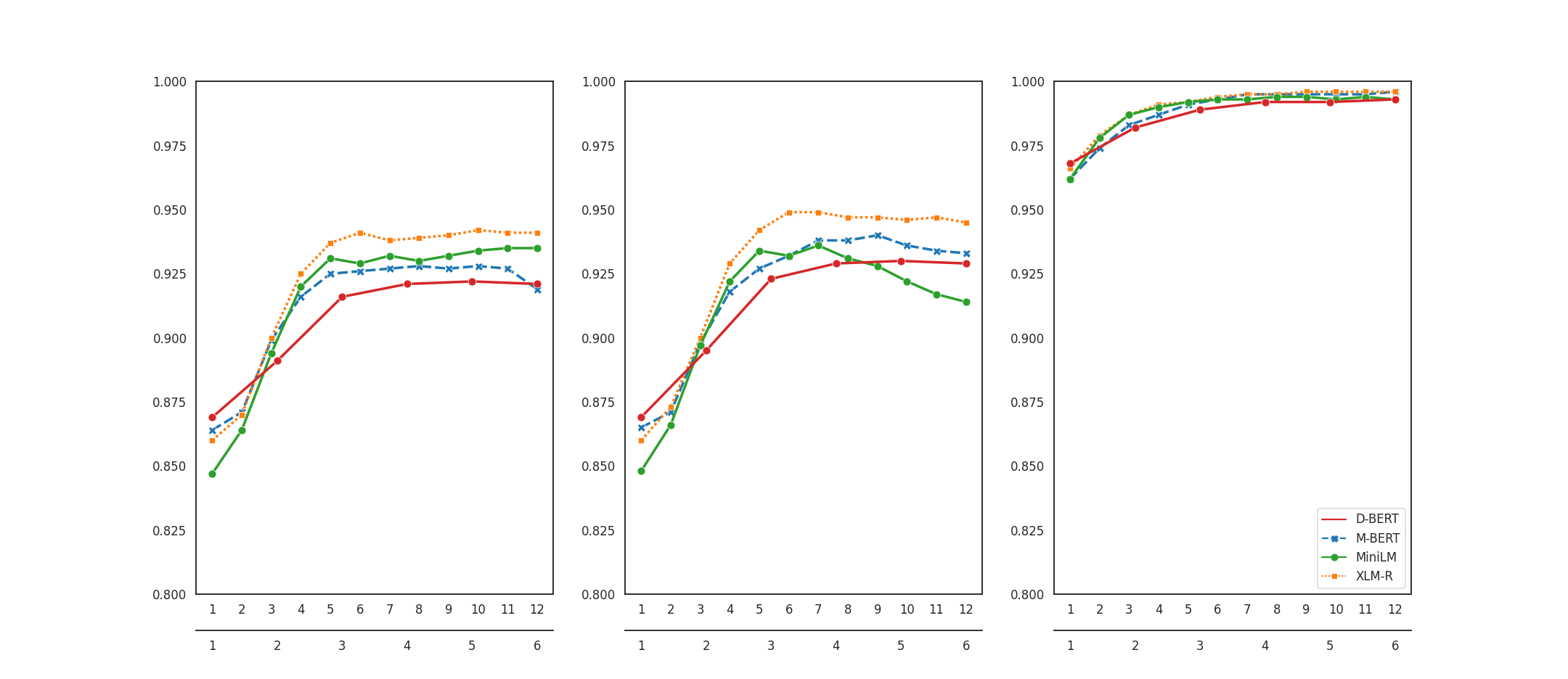}
  \caption{The performance of the probing classifier on \textbf{Morphosyntactic Features} tasks. Left: \textbf{Gender} in German (pre-trained). Middle: \textbf{Gender} in German (fine-tuned). Right: \textbf{Case} in Russian (fine-tuned).}
  \label{morphosynt_feats}
\end{figure*}

\paragraph{Morphosyntactic Features} Tables \ref{tab:feats_pretrained} -- \ref{tab:feats_finetuned} summarize the results of the probing classifier on \textbf{Morphosyntactic Features} tasks for pre-trained and fine-tuned models. Figure \ref{morphosynt_feats} shows a few examples of the model behavior on the tasks. While \textbf{Gender} in German appears to be the most challenging property among the others for both pre-trained and fine-tuned models, \textbf{Case} in Russian is inferred by the models with great confidence.

\paragraph{Masked Token} Tables \ref{tab:mask-nottuned} -- \ref{tab:mask-tuned} outline the performance of the probing classifier on \textbf{Masked Token} tasks.

\begin{figure*}[h]
  \centering
  \includegraphics[width=\textwidth]{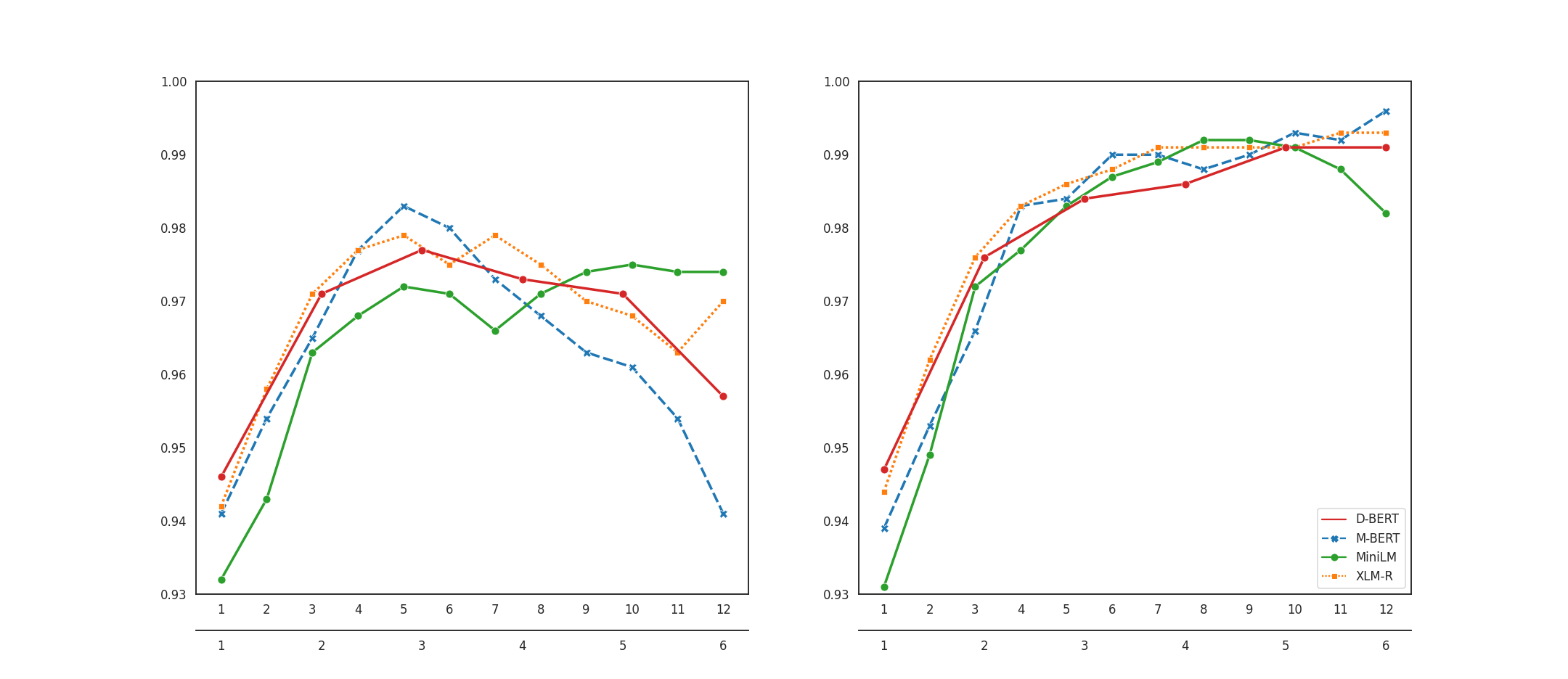}
  \caption{The performance of the probing classifier on \textbf{Article Removal} perturbation task for German. X-axis=Layer index number. Y-axis=Accuracy score. Left: pre-trained models. Right: fine-tuned models.}
  \label{clf_de_article}
\end{figure*}

\begin{figure*}[h]
  \centering
  \includegraphics[width=\textwidth]{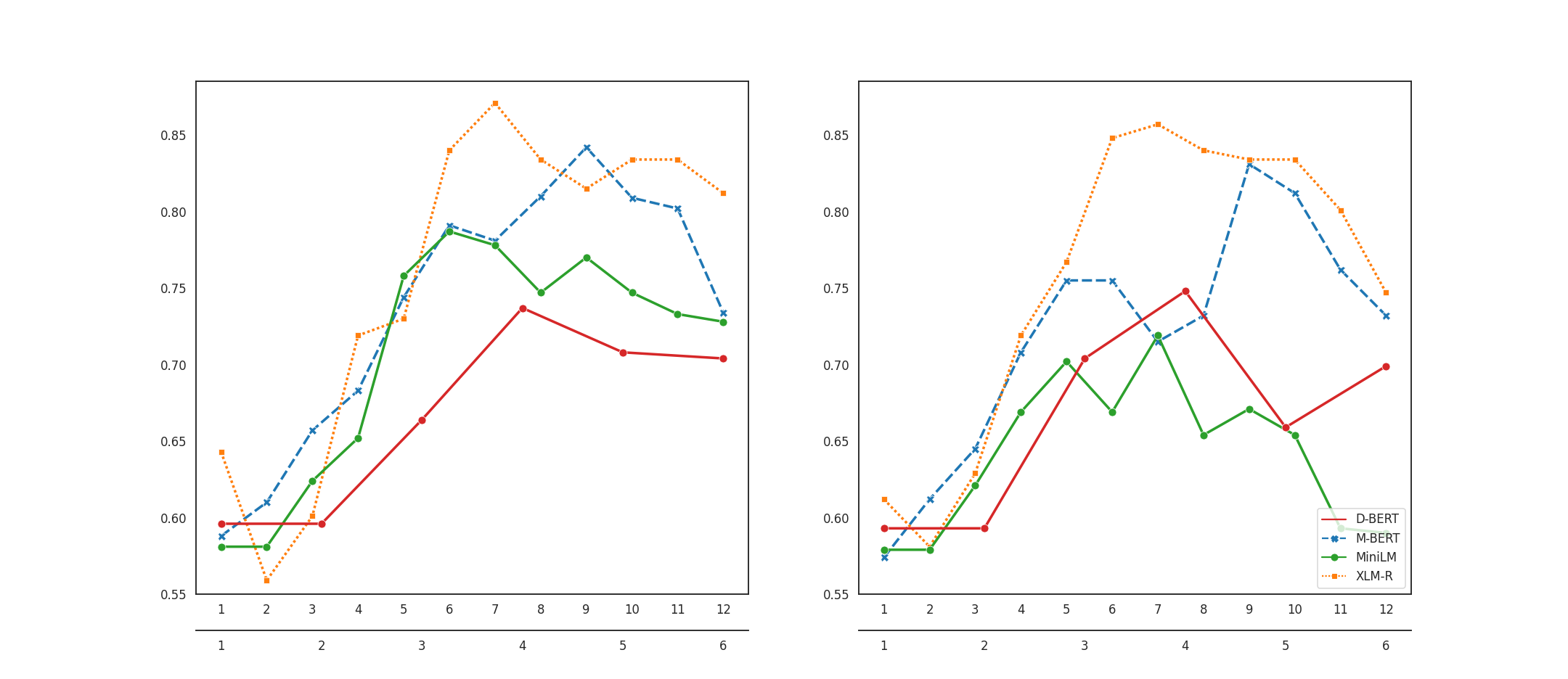}
  \caption{The performance of the probing classifier on \textbf{Predicate Number} perturbation task for French. X-axis=Layer index number. Y-axis=Accuracy score. Left: pre-trained models. Right: fine-tuned models.}
  \label{clf_fr_verb_number}
\end{figure*}

\paragraph{Perturbations} Tables \ref{err-nottuned} -- \ref{err-tuned} present the results of the probing classifier on \textbf{Perturbations} tasks for pre-trained and fine-tuned models. Figures \ref{clf_de_article} -- \ref{clf_fr_verb_number} are the graphical representations of the probing classifier performance on \textbf{Article Removal} task for German, and \textbf{Predicate Number} task for French.

The overall pattern for the syntactic perturbations is that the sensitivity is captured at the lower-to-middle layers $[3-8]$ of pre-trained models. In its turn, the inflectional properties are predominantly distributed at the middle-to-higher layers $[5-12]$ of both pre-trained and fine-tuned models. However, fine-tuned versions may exhibit unpredictable behavior, an example of which we describe below. Figure \ref{clf_de_article} demonstrates the results on \textbf{Article Removal} task for German. While the probing curves of pre-trained models tend to be decaying after reaching their peak at the middle layers, they are confidently increasing towards the output layer after the fine-tuning phase. In contrast, a different behavior is observed on \textbf{Predicate Number} task for French (see Figure \ref{clf_fr_verb_number}). The layers of many fine-tuned models lose their knowledge (\textbf{MiniLM}: $[5-12]$, \textbf{D-BERT}: $[5]$, \textbf{M-BERT}: $[6-11]$, \textbf{XLM-R}: $[7; 11-12]$). 

\begin{table*}[h]\centering

\begin{tabular}{c|c|c|c|c|c}
\toprule
\textbf{Lang}  & \textbf{Probing Task} &  \textbf{D-BERT} & \textbf{MiniLM} &  \textbf{BERT} &  \textbf{XLM-R} \\
\midrule
\multirow{4}{*}{\textbf{De}} 
   & \textbf{Case}   &         0.89 &    \textbf{0.91} &               0.89 &   --- \\
   & \textbf{Gender} &         0.91 &    \textbf{0.92} &               0.91 &  \textbf{0.92} \\
   & \textbf{Number} &         0.93 &    \textbf{0.94} &               0.93 &  \textbf{0.94} \\
   & \textbf{Person} &         0.95 &    \textbf{0.96} &               0.95 &   --- \\
\midrule
\multirow{2}{*}{\textbf{En}} 
   & \textbf{Number} &         0.95 &    \textbf{0.96} &               \textbf{0.96} &  \textbf{0.96} \\
   & \textbf{Person} &         0.98 &    \textbf{0.99} &               0.98 &  \textbf{0.99} \\
\midrule
\multirow{3}{*}{\textbf{Fr}}
   & \textbf{Gender} &         0.92 &    0.92 &               0.92 &  \textbf{0.93} \\
   & \textbf{Number} &         \textbf{0.94} &    0.93 &               \textbf{0.94} &  \textbf{0.94} \\
   & \textbf{Person} &         0.96 &    \textbf{0.97} &               0.96 &  \textbf{0.97} \\
\midrule
\multirow{4}{*}{\textbf{Ru}}
   & \textbf{Case}   &         0.98 &    \textbf{0.99} &               0.98 &  \textbf{0.99} \\
   & \textbf{Gender} &         0.96 &    0.97 &               0.96 &  \textbf{0.98} \\
   & \textbf{Number} &         0.98 &    0.98 &               0.98 &  \textbf{0.99} \\
   & \textbf{Person} &         0.98 &    \textbf{0.99} &               0.98 &  \textbf{0.99} \\
\bottomrule
\end{tabular}

\caption{The results of the probing classifier on \textbf{Morphosyntactic Features} tasks for pre-trained models. The scores are averaged across all layers. Languages: \textbf{Ru}=Russian, \textbf{En}=English, \textbf{De}=German, \textbf{Fr}=French.}\label{tab:feats_pretrained}
\end{table*}

\begin{table*}\centering

\begin{tabular}[h]{c|c|c|c|c|c}
\toprule
\textbf{Lang}  & \textbf{Probing Task} &  \textbf{D-BERT} & \textbf{MiniLM} &  \textbf{BERT} &  \textbf{XLM-R} \\
\midrule
\multirow{4}{*}{\textbf{De}} 
   & \textbf{Case}   &         0.91 &    0.92 &               0.91 &  \textbf{0.93} \\
   & \textbf{Gender} &         0.91 &    0.91 &               0.92 &  \textbf{0.93} \\
   & \textbf{Number} &         0.94 &    0.94 &               0.94 &  \textbf{0.95} \\
   & \textbf{Person} &         0.95 &    0.96 &               0.96 &  \textbf{0.97} \\

\midrule
\multirow{2}{*}{\textbf{En}} 
   & \textbf{Number} &         0.96 &    0.96 &               0.96 &  \textbf{0.97} \\
   & \textbf{Person} &         0.98 &    \textbf{0.99} &               0.98 &  \textbf{0.99} \\

\midrule
\multirow{3}{*}{\textbf{Fr}}
   & \textbf{Gender} &         \textbf{0.93} &    0.92 &               \textbf{0.93} &  \textbf{0.93} \\
   & \textbf{Number} &         0.94 &    0.93 &               \textbf{0.95} &  0.94 \\
   & \textbf{Person} &         0.96 &    \textbf{0.97} &               0.96 &  \textbf{0.97 }\\

\midrule
\multirow{4}{*}{\textbf{Ru}} 
   & \textbf{Case}   &         \textbf{0.99} &    \textbf{0.99} &               \textbf{0.99} &  \textbf{0.99} \\
   & \textbf{Gender} &         0.97 &    0.97 &               0.97 &  \textbf{0.98} \\
   & \textbf{Number} &         \textbf{0.99} &    \textbf{0.99} &               \textbf{0.99} &  \textbf{0.99} \\
   & \textbf{Person} &         \textbf{0.99} &    \textbf{0.99} &               \textbf{0.99} &  \textbf{0.99} \\
\bottomrule
\end{tabular}

\caption{The results of the probing classifier on \textbf{Morphosyntactic Features} tasks for fine-tuned models. The scores are averaged across all layers. Languages: \textbf{Ru}=Russian, \textbf{En}=English, \textbf{De}=German, \textbf{Fr}=French.}\label{tab:feats_finetuned}
\end{table*}

\begin{table*}[h]\centering
\begin{tabular}{c|c|c|c|c|c}\toprule
\textbf{Lang} & \textbf{Probing Task}   &\textbf{D-BERT}  &\textbf{MiniLM} &\textbf{BERT} &\textbf{XLM-R} \\ \midrule

\multirow{2}{*}{\textbf{De}} & \textbf{Gender} & --- & --- & --- & ---\\
 & \textbf{Number} & --- & --- & --- & --- \\
\midrule
\multirow{2}{*}{\textbf{En}} & \textbf{Gender} & 0.52 &0.50 &\textbf{0.53} &0.51 \\
 & \textbf{Number} & 0.66 &0.59 &\textbf{0.68} &0.67 \\
\midrule

\multirow{2}{*}{\textbf{Fr}} & \textbf{Gender} & \textbf{0.72} &0.68 &0.66 &0.69 \\
 & \textbf{Number} & 0.70 &0.65 &0.71 &\textbf{0.73} \\
\midrule

\multirow{4}{*}{\textbf{Ru}} & \textbf{Case} & 0.67 &0.61 & 0.74 &\textbf{0.78} \\
 & \textbf{Gender} & 0.68 & 0.67 & 0.67 & \textbf{0.73} \\
 & \textbf{Number} & 0.67 & 0.63 & 0.71 & \textbf{0.75} \\
 & \textbf{Person} & 0.62 & 0.51 & 0.59 & \textbf{0.69} \\

\bottomrule
\end{tabular}
\caption{The results of the probing classifier on \textbf{Masked Tokens} tasks for pre-trained models. The scores are averaged across all layers. Languages: \textbf{Ru}=Russian, \textbf{En}=English, \textbf{De}=German, \textbf{Fr}=French.}\label{tab:mask-nottuned}
\end{table*}

\begin{table*}[h]\centering
\begin{tabular}{c|c|c|c|c|c}\toprule
\textbf{Lang} & \textbf{Probing Task}   &\textbf{D-BERT}  &\textbf{MiniLM} &\textbf{BERT} &\textbf{XLM-R} \\ \midrule

\multirow{2}{*}{\textbf{De}} & \textbf{Gender} & --- & --- & --- & ---\\
 & \textbf{Number} & --- & --- & --- & --- \\
\midrule
\multirow{2}{*}{\textbf{En}} & \textbf{Gender} & 0.51 & 0.51 & 0.51 & \textbf{0.52} \\
 & \textbf{Number} & 0.64 & 0.61 & 0.64 & \textbf{0.66}\\
\midrule

\multirow{2}{*}{\textbf{Fr}} & \textbf{Gender} &  0.67 & 0.60 & \textbf{0.69} & 0.62 \\
 & \textbf{Number} & 0.63 & 0.65 & \textbf{0.69} & 0.59\\
\midrule

\multirow{4}{*}{\textbf{Ru}} & \textbf{Case} & 0.67 & 0.62 & 0.67 & \textbf{0.70} \\
 & \textbf{Gender} & 0.67 & 0.62 & 0.50 & \textbf{0.68}\\
 & \textbf{Number} & \textbf{0.66} & 0.60 & 0.5 & 0.63\\
 & \textbf{Person}& 0.52 & 0.52 & 0.53 & \textbf{0.55}\\
 \bottomrule
\end{tabular}
\caption{The results of the probing classifier on \textbf{Masked Tokens} tasks for fine-tuned models. The scores are averaged across all layers. Languages: \textbf{Ru}=Russian, \textbf{En}=English, \textbf{De}=German, \textbf{Fr}=French.}\label{tab:mask-tuned}
\end{table*}

\begin{table*}[h]\centering

\begin{tabular}{c|c|c|c|c|c}\toprule
\textbf{Lang} & \textbf{Probing Task}   &\textbf{D-BERT}  &\textbf{MiniLM} &\textbf{BERT} &\textbf{XLM-R} \\
\midrule

\multirow{4}{*}{\textbf{De}} & \textbf{Article Removal} &\textbf{0.97} &0.96 &0.96 &\textbf{0.97} \\
& \textbf{Deixis Word Number} &0.65 &0.63 &0.66 &\textbf{0.73} \\
& \textbf{Subject Number} &0.6 &0.66 &0.68 &\textbf{0.72} \\
& \textbf{Predicate Number} &0.67 &0.67 &0.72 &\textbf{0.77} \\
\midrule
\multirow{4}{*}{\textbf{En}} & \textbf{Article Removal} & \textbf{0.98} &0.97 &0.97 &0.97 \\
& \textbf{Stop-words Removal} &\textbf{0.99} &\textbf{0.99} &0.98 &\textbf{0.99} \\
& \textbf{Subject Number} &\textbf{0.53} &\textbf{0.53} &0.51 &0.51 \\
& \textbf{Predicate Number} &0.51 &0.53 &0.52 &\textbf{0.59} \\
\midrule
\multirow{3}{*}{\textbf{Fr}} & \textbf{Article Removal} &0.96 &0.96 &0.95 &\textbf{0.97} \\
& \textbf{Subject Number} &0.63 &0.65 &\textbf{0.71} &\textbf{0.71} \\
& \textbf{Predicate number} &0.67 &0.71 &0.74 &\textbf{0.76 }\\
\midrule
\multirow{6}{*}{\textbf{Ru}}  &  \textbf{Stop-words Removal} &0.95 &\textbf{0.96} &0.94 &\textbf{0.96} \\
& \textbf{Subject Case} &0.72 &0.75 &0.77 &\textbf{0.82} \\
& \textbf{Subject Number} &0.63 &0.68 &0.7 &\textbf{0.76} \\
& \textbf{Predicate Gender} &0.63 &0.64 &0.67 &\textbf{0.71 }\\
& \textbf{Predicate Number} &0.64 &0.67 &0.71 &\textbf{0.75} \\
& \textbf{Predicate Person} &0.77 &0.8 &0.81 &\textbf{0.86} \\
\bottomrule
\end{tabular}

\caption{The results of the probing classifier on \textbf{Perturbations} tasks for pre-trained models. The scores are averaged across all layers. Languages: \textbf{Ru}=Russian, \textbf{En}=English, \textbf{De}=German, \textbf{Fr}=French.}\label{err-nottuned}
\end{table*}

\begin{table*}[h]\centering
\begin{tabular}{c|c|c|c|c|c}\toprule
\textbf{Lang} & \textbf{Probing Task}   &\textbf{D-BERT}  &\textbf{MiniLM} &\textbf{BERT} &\textbf{XLM-R} \\
\midrule
\multirow{4}{*}{\textbf{De}} & \textbf{Article Removal} &\textbf{0.98} &\textbf{0.98} &\textbf{0.98} &\textbf{0.98} \\
& \textbf{Deixis Word Number} &0.63 &0.63 &0.69 &\textbf{0.72} \\
& \textbf{Subject Number} &0.62 &0.62 &0.68 &\textbf{0.71} \\
& \textbf{Predicate Number} &0.67 &0.64 &0.7 &\textbf{0.75} \\
\midrule
\multirow{4}{*}{\textbf{En}} & \textbf{Article Removal} &\textbf{0.99} &0.98 &\textbf{0.99} &\textbf{0.99} \\
& \textbf{Stop-words Removal} &\textbf{0.99} &\textbf{0.99} &\textbf{0.99} &\textbf{0.99} \\
& \textbf{Subject Number} &0.54 &0.52 &\textbf{0.57} &0.55 \\
& \textbf{Predicate Number} &0.51 &0.52 &0.54 &\textbf{0.6} \\
\midrule
\multirow{3}{*}{\textbf{Fr}} & \textbf{Article} &0.97 &0.96 &0.96 &\textbf{0.98} \\
& \textbf{Subject Number} &0.65 &0.67 &0.73 &\textbf{0.76} \\
& \textbf{Predicate Number} &0.67 &0.64 &0.72 &\textbf{0.76} \\
\midrule

\multirow{6}{*}{\textbf{Ru}}  & \textbf{Stop-words Removal} &0.96 &0.95 &0.96 &\textbf{0.97} \\
& \textbf{Subject Case} &0.75 &0.75 &0.79 &\textbf{0.84} \\
& \textbf{Subject Number} &0.65 &0.65 &0.72 &\textbf{0.77} \\
& \textbf{Predicate Gender} &0.63 &0.62 &0.67 &\textbf{0.71} \\
& \textbf{Predicate Number} &0.62 &0.62 &0.7 &\textbf{0.75} \\
& \textbf{Predicate Person} &0.79 &0.8 &0.8 &\textbf{0.85} \\
\bottomrule
\end{tabular}

\caption{The results of the probing classifier on \textbf{Perturbations} tasks for fine-tuned models. The scores are averaged across all layers. Languages: \textbf{Ru}=Russian, \textbf{En}=English, \textbf{De}=German, \textbf{Fr}=French.}\label{err-tuned}
\end{table*}

\clearpage

\section{Individual Neuron Analysis}
\label{app:individual_neuron_an}
\begin{figure*}[h]
  \centering
  \includegraphics[width=.9\textwidth]{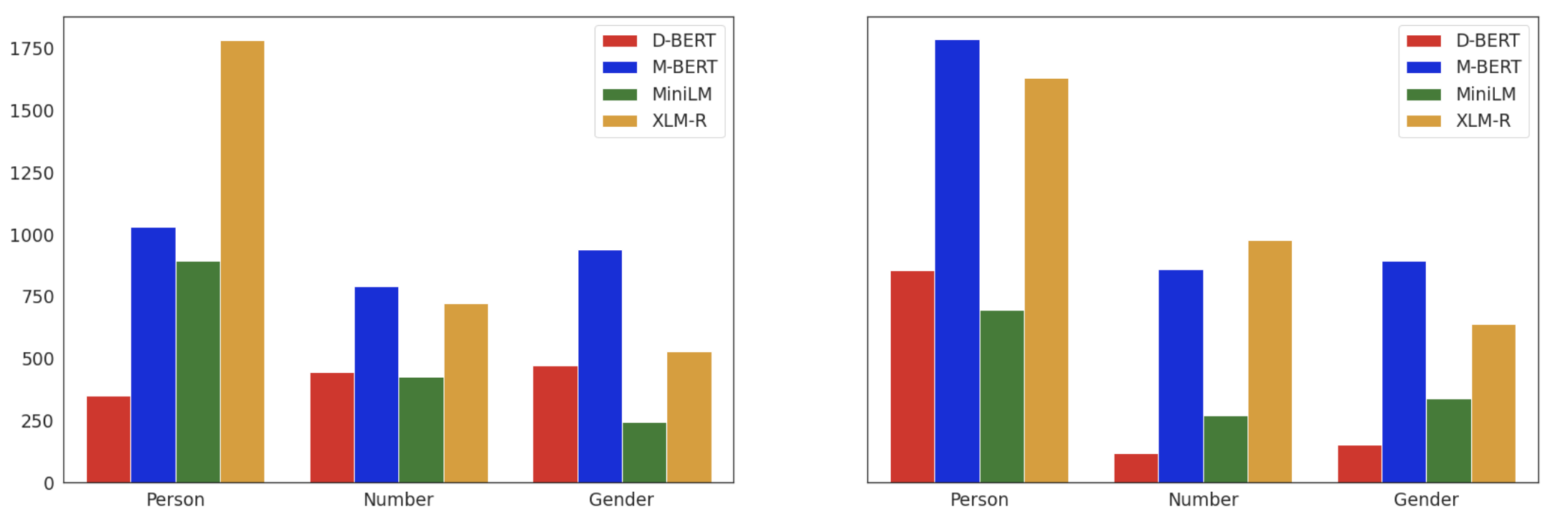}
  \caption{Number of neurons per each property for French. Y-axis=Number of neurons. Left: pre-trained models. Right: fine-tuned models.}
  \label{french_prop_wise}
  

  \centering
  \includegraphics[width=.92\textwidth]{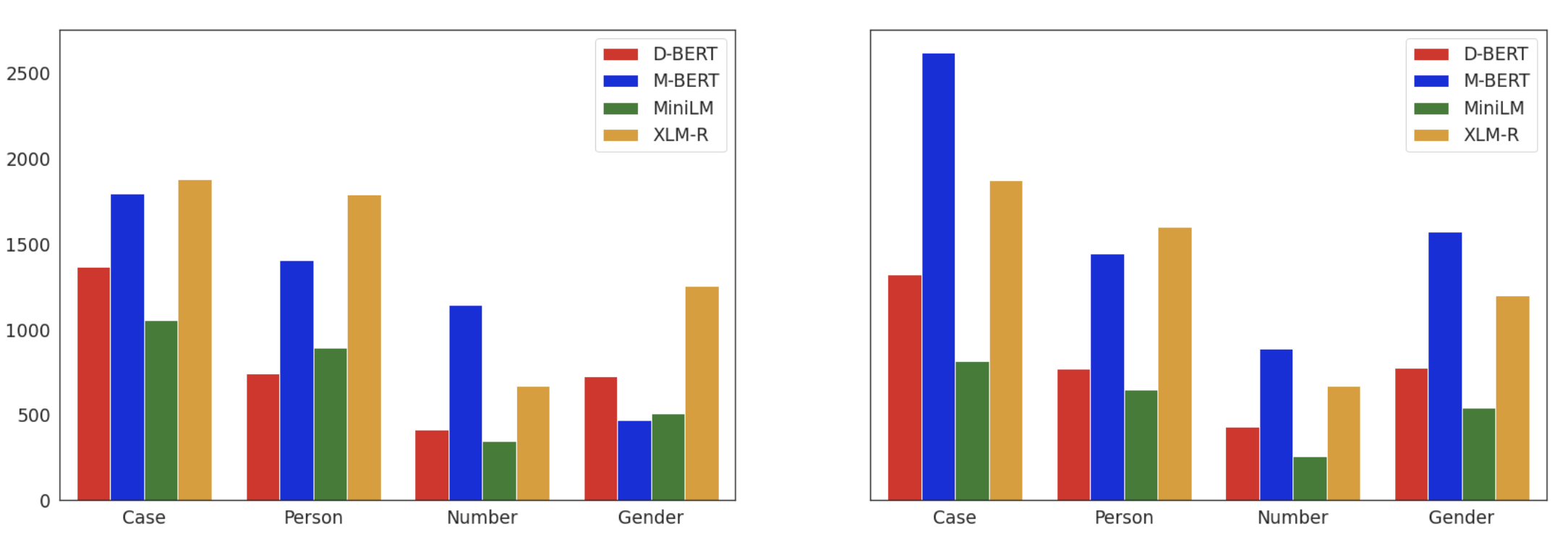}
  \caption{Number of neurons per each property for Russian. Y-axis=Number of neurons. Left: pre-trained models. Right: fine-tuned models.}
  \label{russian_prop_wise}


  \centering
  \includegraphics[width=.92\textwidth]{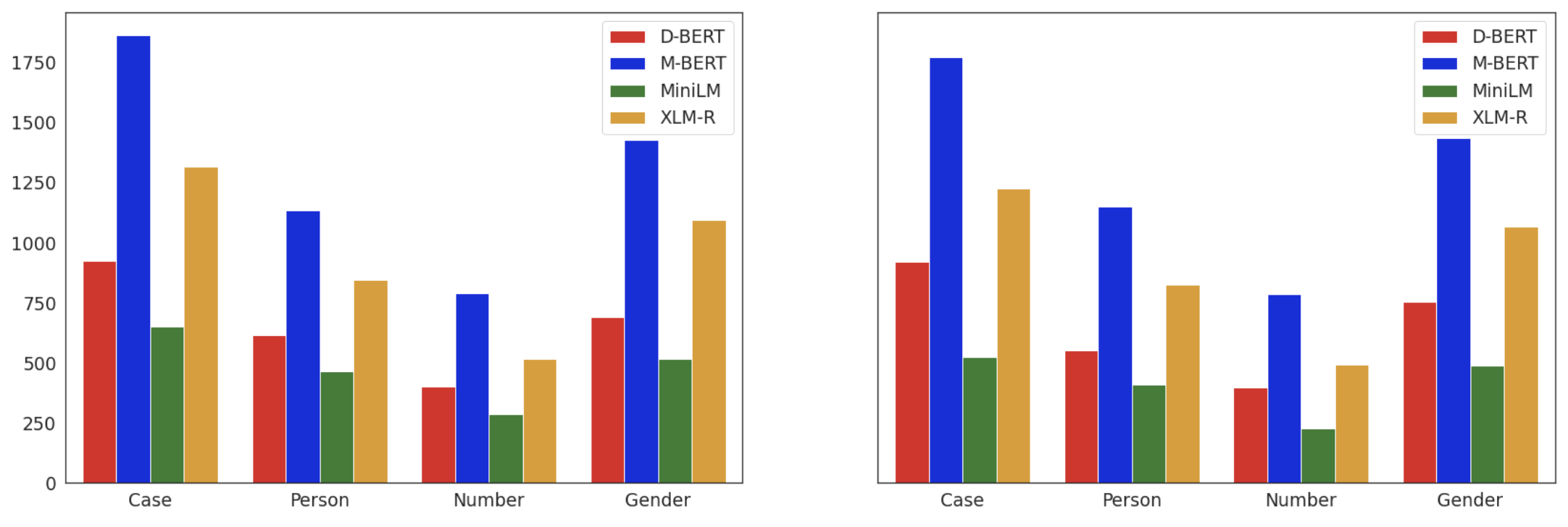}
  \caption{Number of neurons per each property for German. Y-axis=Number of neurons. Left: pre-trained models. Right: fine-tuned models.}
  \label{german_prop_wise}
\end{figure*}

\begin{figure*}[h]
  \centering
  \includegraphics[width=.92\textwidth]{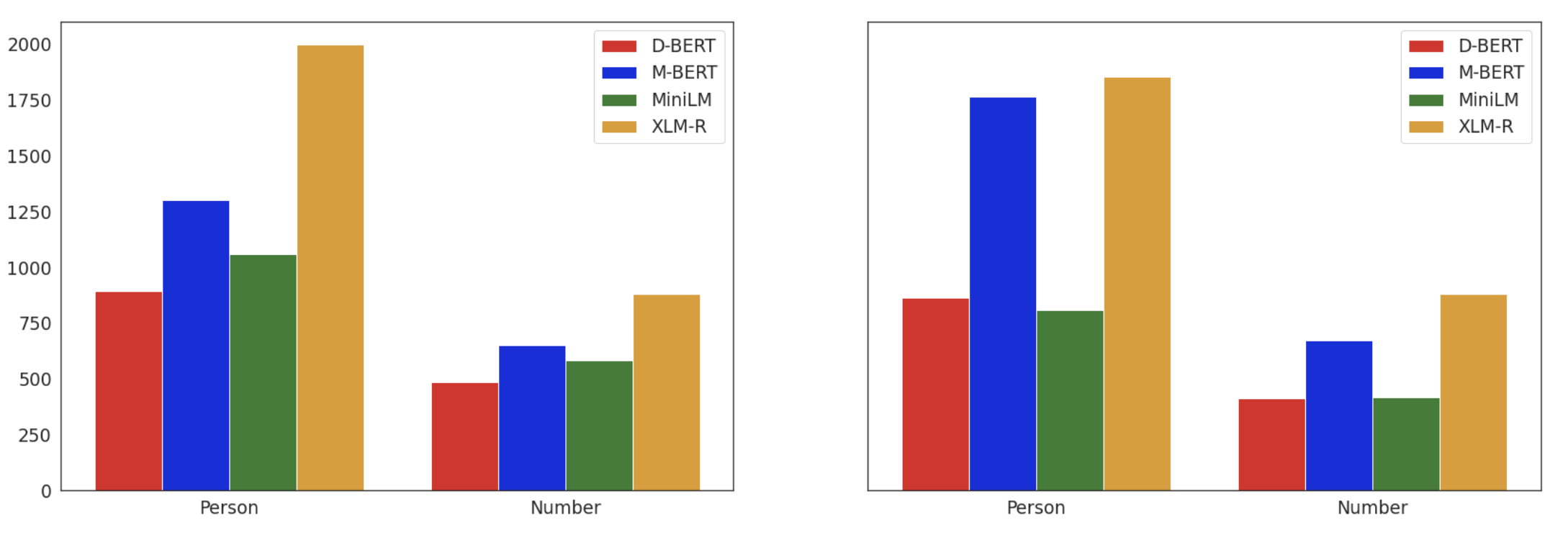}
  \caption{Number of neurons per each property for English. Y-axis=Number of neurons. Left: pre-trained models. Right: fine-tuned models.}
  \label{english_prop_wise}
\end{figure*}

\paragraph{Property-wise analysis} Figures \ref{french_prop_wise} -- \ref{english_prop_wise} depict property-wise neuron distribution for French, Russian, German and English.

\begin{figure*}[h]
  \centering
  \includegraphics[width=.9\textwidth]{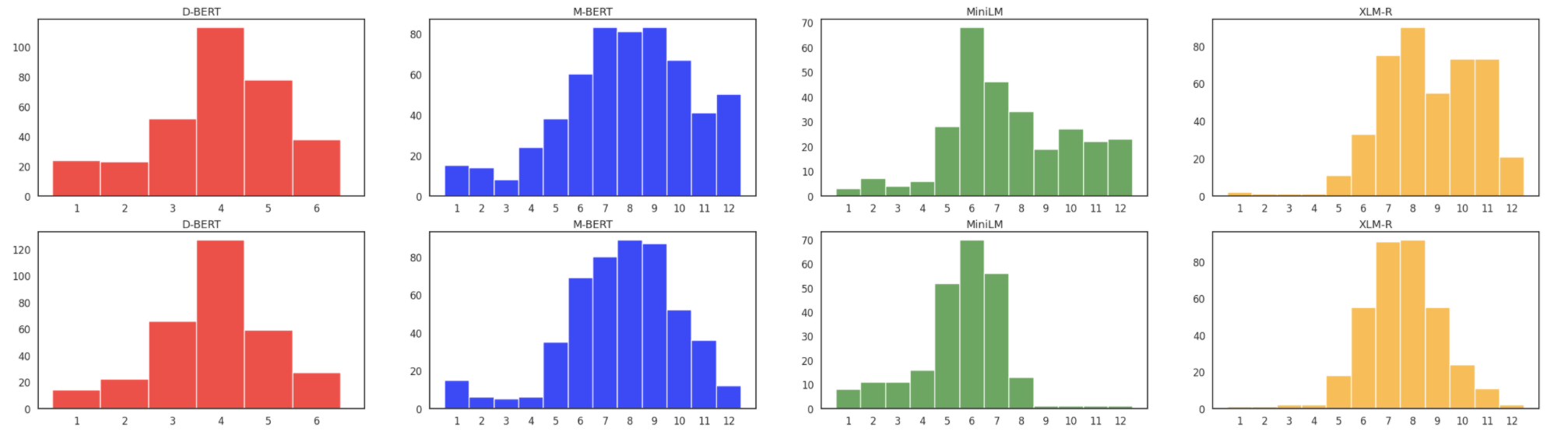}
  \caption{The distribution of top neurons over \textbf{Subject Number} perturbation task for each model (\textbf{Russian}). X-axis=Layer index number. Y-axis=Number of neurons. Top: pre-trained models. Bottom: fine-tuned models.}
  \label{ru_subj_number_distribution}
  \vspace{1.5em}

  \centering
  \includegraphics[width=.9\textwidth]{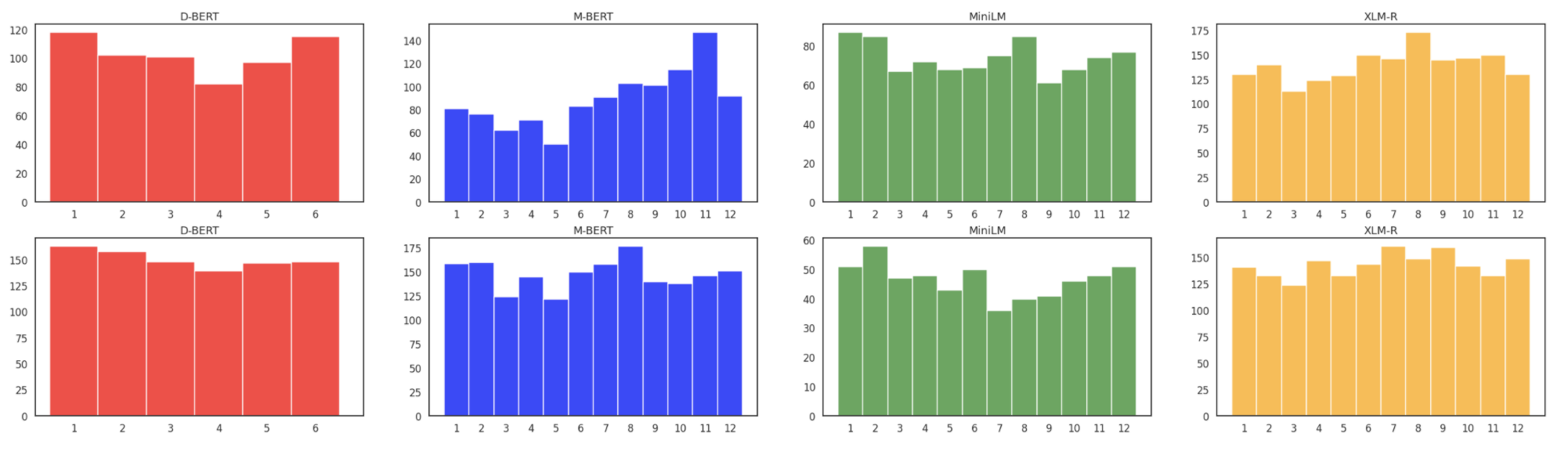}
  \caption{The distribution of top neurons over \textbf{Subject Number} perturbation task for each model (\textbf{English}). X-axis=Layer index number. Y-axis=Number of neurons. Top: pre-trained models. Bottom: fine-tuned models.}
  \label{en_subj_number_distribution}
  \vspace{1.5em}

  \centering
  \includegraphics[width=.9\textwidth]{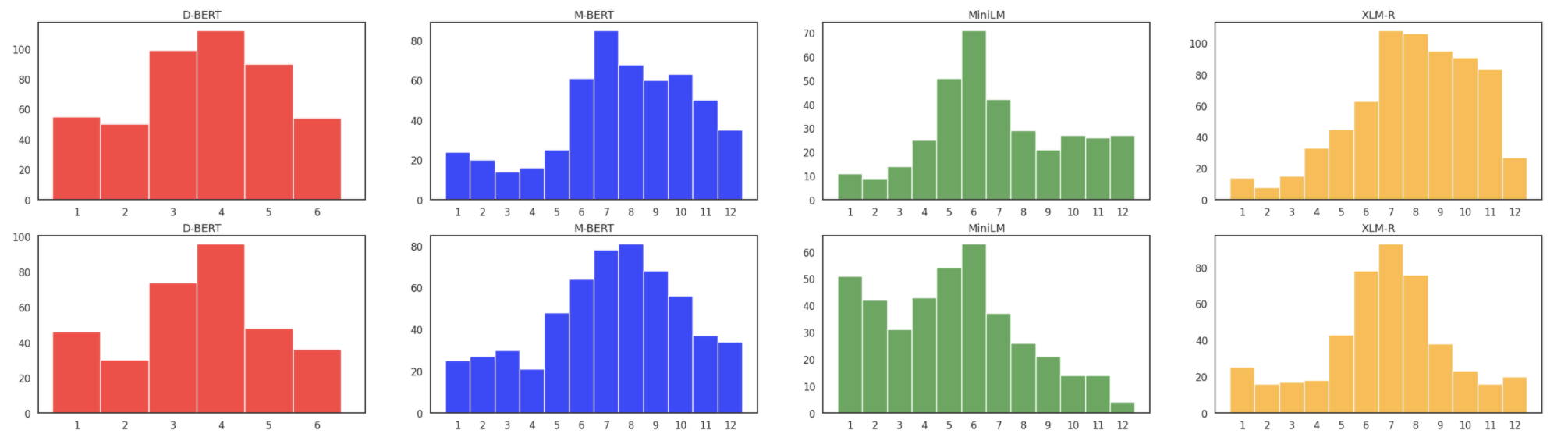}
  \caption{The distribution of top neurons over \textbf{Subject Number} perturbation task for each model (\textbf{French}). X-axis=Layer index number. Y-axis=Number of neurons. Top: pre-trained models. Bottom: fine-tuned models.}
  \label{fr_subj_number_distribution}
\end{figure*}

\paragraph{Layer-wise analysis}
Figures \ref{ru_subj_number_distribution} -- \ref{fr_subj_number_distribution} demonstrate the results of the individual neuron analysis on \textbf{Subject Number} perturbation task for Russian, English and French.

\clearpage

\section{POS-Tagging Performance}
\label{pos_perf}

Tables \ref{pos-de} -- \ref{pos-ru} describe the results of the fine-tuning on POS-tagging task for each language. 

\begin{table*}[ht!]

\centering

\begin{tabular}{c|c|c|c|c}\toprule
\textbf{Model / Metric} &\textbf{Accuracy} &\textbf{F1} &\textbf{Precision} &\textbf{Recall}  \\\midrule
M-BERT &\textbf{0.98} &\textbf{0.98}  &\textbf{0.98} &\textbf{0.98} \\
DistilBERT &\textbf{0.98} &\textbf{0.98} &\textbf{0.98} &\textbf{0.98} \\
MiniLM &\textbf{0.98} &\textbf{0.98} &\textbf{0.98} &\textbf{0.98} \\
XLM-R &\textbf{0.98} &\textbf{0.98} &\textbf{0.98} &\textbf{0.98} \\
\bottomrule
\end{tabular}
\caption{Metrics of the models fine-tuned for POS-tagging task for German.}\label{pos-de}
\vspace{2em}


\begin{tabular}{c|c|c|c|c}\toprule
\textbf{Model / Metric} &\textbf{Accuracy} &\textbf{F1} &\textbf{Precision} &\textbf{Recall}  \\\midrule
M-BERT &\textbf{0.96} &0.95 &0.95 &0.95 \\
DistilBERT &\textbf{0.95} &0.94 &0.94 &0.94 \\
MiniLM &\textbf{0.95} &0.94 &0.94 &0.94 \\
XLM-R &\textbf{0.96} &\textbf{0.96} &\textbf{0.96} &\textbf{0.96} \\
\bottomrule
\end{tabular}
\caption{Metrics of the models fine-tuned for POS-tagging task for English.}\label{pos-en}
\vspace{2em}


\begin{tabular}{c|c|c|c|c}\toprule
\textbf{Model / Metric} &\textbf{Accuracy} &\textbf{F1} &\textbf{Precision} &\textbf{Recall}  \\\midrule
M-BERT &\textbf{0.98} &0.97 &0.97 &0.97 \\
DistilBERT &\textbf{0.97} &\textbf{0.97}  &\textbf{0.97} &\textbf{0.97} \\
MiniLM &\textbf{0.97} &0.96 &0.96 &0.96 \\
XLM-R &\textbf{0.98} &\textbf{0.98} &\textbf{0.98} &\textbf{0.98} \\
\bottomrule
\end{tabular}
\caption{Metrics of the models fine-tuned for POS-tagging task for French.}\label{pos-fr}
\vspace{2em}


\begin{tabular}{c|c|c|c|c}
\toprule
\textbf{Model / Metric} &\textbf{Accuracy} &\textbf{F1} &\textbf{Precision} &\textbf{Recall} \\\midrule
M-BERT &\textbf{0.99} &\textbf{0.99}  &\textbf{0.99} &\textbf{0.99} \\
DistilBERT &\textbf{0.99} &\textbf{0.99} &\textbf{0.99} &\textbf{0.99} \\
MiniLM &\textbf{0.99} &0.98  &0.98 &0.98 \\
XLM-R &\textbf{0.99} &\textbf{0.99}  &\textbf{0.99} &\textbf{0.99} \\
\bottomrule
\end{tabular}
\caption{Metrics of the models fine-tuned for POS-tagging task for Russian.}\label{pos-ru}
\end{table*}

\end{document}